\documentclass[10pt,journal,compsoc]{IEEEtran}

\usepackage{cite}
\usepackage{url}
\usepackage{ragged2e}
\usepackage{epsfig}
\usepackage{graphicx}
\usepackage{amsmath,amssymb} 
\usepackage{subfigure}
\usepackage{algorithm}
\usepackage{algorithmicx}
\usepackage{algpseudocode}
\usepackage{graphics}
\usepackage{threeparttable}
\usepackage{color}
\usepackage[normalem]{ulem}
\usepackage{multirow}
\usepackage{float}
\usepackage{amsfonts}
\usepackage{bm}
\usepackage{array}
\usepackage[table]{xcolor}
\usepackage{colortbl}
\usepackage{pifont}
\usepackage{diagbox}
\usepackage{rotating}
\usepackage{booktabs}
\usepackage{overpic}
\usepackage{textcomp}
\usepackage{contour}
\usepackage{enumitem}
\usepackage[colorlinks=true]{hyperref}
\usepackage{booktabs}

\RequirePackage{silence}
\definecolor{bblue}{rgb}{0,150,230}
\definecolor{mygray}{gray}{.92}

\newcommand{\secref}[1]{Section \ref{#1}}

\graphicspath{{./Imgs/}}
\DeclareGraphicsExtensions{.jpg,.pdf,.png}

\makeatletter
\def\ps@IEEEtitlepagestyle{%
  \def\@oddfoot{\mycopyrightnotice}%
  \def\@evenfoot{}%
}
\def\mycopyrightnotice{%
  {\hfill \scriptsize {This work has been submitted to the IEEE for possible publication.
  Copyright may be transferred without notice, after which this version may no longer be accessible.}\hfill}

}
\makeatother

\begin{document}
\title{Taking an Emotional Look at Video Paragraph Captioning}
\author{Qinyu~Li,
        Tengpeng~Li,
        Hanli~Wang,~\IEEEmembership{Senior Member,~IEEE,}
        Chang~Wen~Chen,~\IEEEmembership{Fellow,~IEEE}

        \thanks{
         Qinyu Li and Tengpeng~Li are co-first authors.}

         \thanks{Corresponding author: Hanli~Wang.}

         \thanks{{Q.~Li, T.~Li and H.~Wang are with the Department of Computer Science \& Technology, Key Laboratory of Embedded System and Service Computing (Ministry of Education), Tongji University, Shanghai 200092, P. R. China, and with Frontiers Science Center for Intelligent Autonomous Systems, Shanghai 201210 (e-mail: qinyu.li@tongji.edu.cn, ltpfor1225@tongji.edu.cn, hanliwang@tongji.edu.cn).}}

         \thanks{{C. W. Chen is with the Department of Computing, Hong Kong Polytechnic University, Hung Hom, Kowloon, Hong Kong SAR, China (e-mail: Changwen.chen@polyu.edu.hk).}}
        }


\IEEEtitleabstractindextext{
\begin{abstract}
\justifying
Translating visual data into natural language is essential for machines to understand the world and interact with humans. In this work, a comprehensive study is conducted on video paragraph captioning, which is a developing and challenging task evolved from video captioning, with the goal to generate paragraph-level descriptions for a given video. However, current researches mainly focus on detecting objective facts (\textit{e.g.}, objects, actions and events), ignoring the needs to establish the logical associations between sentences and to discover more accurate emotions related to video contents. Such a problem of ignorance impairs fluent and abundant expressions of predicted captions, which are far below human language standards. To solve this problem, we propose to construct a large-scale emotion and logic driven multilingual dataset for this task. This dataset is named EMVPC (standing for ``Emotional Video Paragraph Captioning'') and contains 53 widely-used emotions in daily life, 376 common scenes corresponding to these emotions, 10,291 high-quality videos and 20,582 elaborated paragraph captions with English and Chinese versions. Relevant emotion categories, scene labels, emotion word labels and logic word labels are also provided in this new dataset. The proposed EMVPC dataset intends to provide full-fledged video paragraph captioning in terms of rich emotions, coherent logic and elaborate expressions, which can also benefit other tasks in vision-language fields. Furthermore, a comprehensive study is conducted through experiments on existing benchmark video paragraph captioning datasets (Charades Captions and ActivityNet Captions) and the proposed EMVPC. The state-of-the-art schemes from different visual captioning tasks are compared in terms of 15 popular metrics, and their detailed objective as well as subjective results are summarized. Finally, remaining problems and future directions of video paragraph captioning are also discussed. The unique perspective of this work is expected to boost further development in video paragraph captioning research.
\end{abstract}
}

\maketitle
\IEEEdisplaynontitleabstractindextext

\IEEEpeerreviewmaketitle

\IEEEraisesectionheading{\section{Introduction}\label{sec:introduction}}

Automatically describing video content with natural language is a fundamental and challenging task, and has attracted rapidly growing attention in computer vision and natural language processing~(NLP) communities. It has promising applications in human-robot interaction~\cite{Cascianelli2018full,das2017visual}, video retrieval~\cite{xu2019multilevel,bansal2019visual} and assisting the visually-impaired~\cite{wu2017automatic,bennett2018how}. The long-standing problem in computer vision and NLP is the semantic gap between low-level visual data and high-level abstract knowledge. To bridge the semantic gap, dominant video captioning models~\cite{venugopalan2014translating,yao2015describing,zhang2020object,pan2020spatio} have been proposed to generate a single sentence to describe video content which primarily covers one main event happened in a short duration. However, the videos in real-world situations are usually untrimmed and contain abundant events along the temporal dimension. In general, information extracted from long videos cannot be fully expressed by a single sentence. Therefore, video paragraph captioning methods~\cite{park2019adversarial,lei2020mart,song2021towards} have been reported to explore multiple events in video content and depict these events with paragraph-level descriptions.

In video paragraph captioning, it is essential to generate logical, abundant and coherent expressions based on video contents, and meaningful contextual information needs to be deduced for language generation. Generally, the expression abilities of existing video paragraph captioning methods are limited in two aspects: (1) learning representation from given contextual information and (2) ensuring logical structure in story-oriented descriptions. On one hand, in addition to extract the 2D appearance features~(\textit{e.g.}, ResNeXt101~\cite{park2019adversarial}, ResNet200~\cite{song2021towards}), most video captioning methods also concentrate on employing other features, such as motion features~\cite{zhou2018end,pan2020spatio,song2021towards}, audio features~\cite{BMT_Iashin_2020} and 3D features~\cite{luo2020univl}. However, these features are all extracted according to video contents and still have a semantic gap with textual representation, leading to inaccurate translation. In order to extract textual representation directly, several recent methods~\cite{yang2020hierarchical,zhong2020comprehensive,zhang2021open} introduced external knowledge~(\textit{e.g.}, scene graph~\cite{xu2017scene}, video retrieval content~\cite{gabeur2020multi}) into text generation module. Nevertheless, these additional contexts are learned by post-processing visual information instead of utilizing intrinsic annotated captions directly.

On the other hand, several effective models~\cite{lei2020mart,song2021towards} were proposed to enhance logical expression of the generated paragraph. The recurrent memory mechanism~\cite{lei2020mart} aimed at producing a generalized memory state to guide the next sentence generation, and the dynamic video memory module~\cite{song2021towards} was designed to enhance logical construction in the temporal dimension. Consequently, the logical structure containing contextual information is crucial to output coherent sentences. However, these two methods~\cite{lei2020mart,song2021towards} can only establish superficial logical association between sentences. Due to deep dependency on the characteristics of training data, they still cannot produce human-like language with logicality and coherence. For existing video paragraph captioning datasets, logical association between sentences and informative representation in captions are usually overlooked.

In addition, emotion expression is essential for visual captioning from basic factual description to high-level emotional narration. Benefitting from deep learning techniques, the researches in visual sentiment analysis~\cite{xu2018heterogeneous,zhao2017microblog,yadav2019xra} have advanced rapidly with great success. However, most visual captioning approaches neglect emotion expression and only focus on objective factual description, resulting in plain, rigid and bored descriptions. Facing this challenge, the emotion captioning datasets FlickrStyle10K~\cite{gan2017stylenet} and EmVidCap~\cite{wang2021emotion} were constructed to enrich emotion expression for the generated sentences.
Specifically, FlickrStyle10K~\cite{gan2017stylenet} contains 10K Flickr images with stylized captions (\textit{i.e.}, humour and romance). The style factors can be disentangled from sentences and incorporated to generate captions with different styles. Nonetheless, this dataset only covers two specific styles, restricting the model's capacity to produce adequate expressions. EmVidCap~\cite{wang2021emotion} consists of 1,897 videos covered with 34 common emotion categories, and emotion words are embedded into most descriptive sentences. A fact stream and an emotion stream are designed to generate emotion expressions for videos. Emotion word accuracy and emotion sentence accuracy are used as the emotion evaluation metrics. Although this method of constructing an emotion-embedding dataset is easy to realize, it inevitably lacks logical coherence. In summary, FlickrStyle10K~\cite{gan2017stylenet} and EmVidCap~\cite{wang2021emotion} contain small-scale images or videos, few emotion categories and short sentence descriptions. They are not suitable for designing new schemes that can significantly improve emotional and logical expression for paragraph captioning.

\begin{figure*}[tb]
\begin{center}
\begin{tabular}{c}
\includegraphics[width=0.9\linewidth]{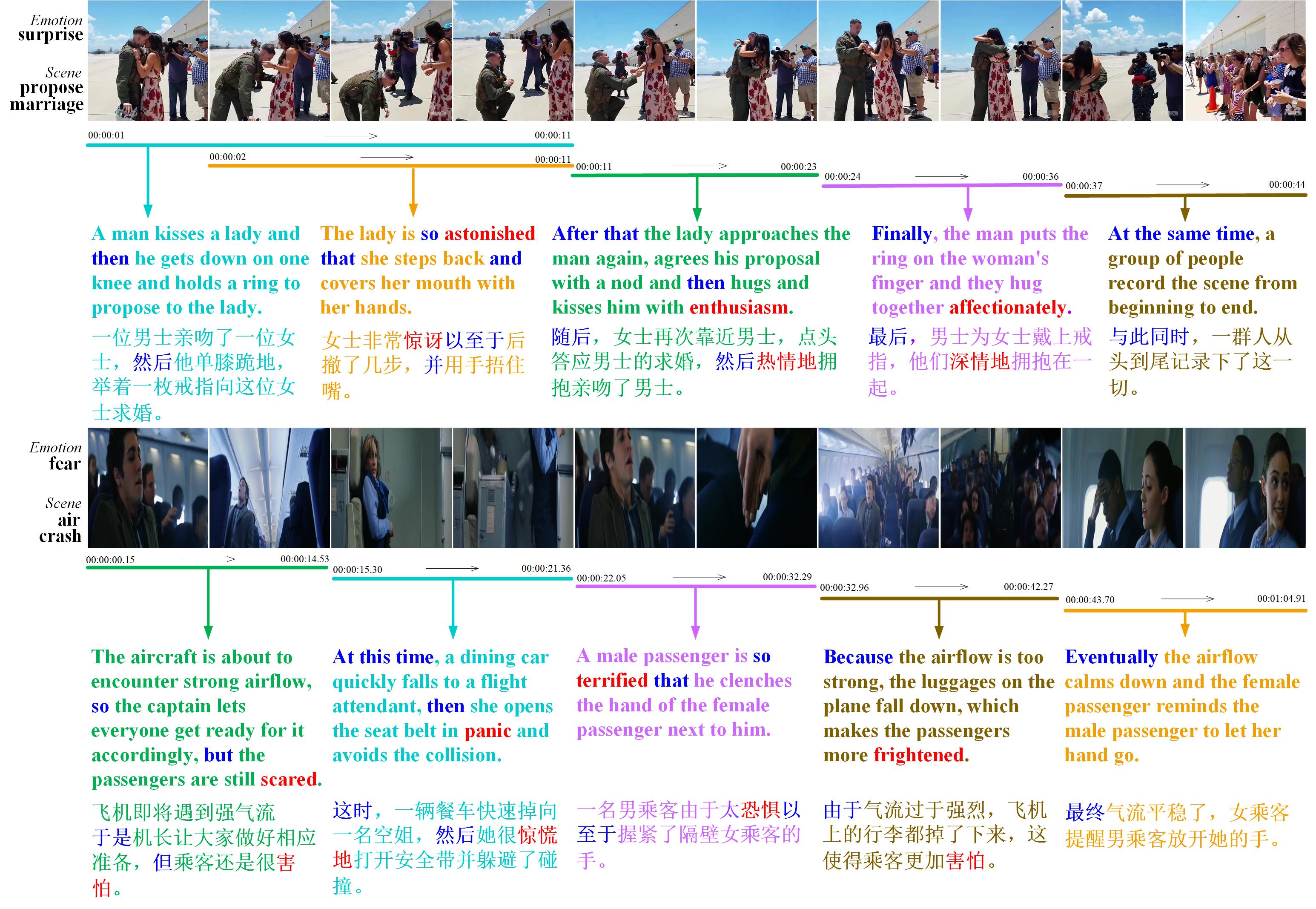}
\end{tabular}
\end{center}
   \caption{ Video samples of the proposed EMVPC dataset. It has rich information, \textit{i.e.}, emotion categories, scene labels, emotion words, logic words, elaborate expressions and multilingual. The segment of each video has a time stamp label and is annotated with one sentence. The emotional and logical representations of the annotated caption are highlighted in \textcolor[rgb]{1.00,0.00,0.00}{red} and \textcolor[rgb]{0.00,0.00,1.00}{blue}, respectively.}
\label{fig:comparison-visual-fig1}
\end{figure*}

To address the aforementioned problems, a large-scale dataset named EMVPC (Emotional Video Paragraph Captioning) is constructed for conducting research in video paragraph captioning. This dataset contains 10,291 unique videos and 20,582 elaborated captions. In particular, this dataset consists of 53 human emotions and 376 real-world scenes, in which each video is supplied with one English paragraph caption, one Chinese paragraph caption, emotion category, scene label, topic-aware emotion words and logic words. The goal of constructing EMVPC is to provide a large-scale corpus with rich emotions, coherent logic and elaborate expressions, promoting possible paradigm shifting research in video paragraph captioning from monotonous descriptions to rich and coherent expressions. To this end, all annotating volunteers were trained to comprehend video content from different perspectives. Specifically, they not only provide objective statements about the superficial information obtained from the given videos (\textit{e.g.}, objects, characters, environments, events), but also describe the deep understanding of multiple complex scenarios (\textit{e.g.}, summarizing the topic-aware emotion, clarifying the logical relations between different events, inferring the potential activity out of the video). Different from the superficial factual captions in the existing video captioning datasets~\cite{xu2016msrvtt,krishna2017dense,wang2018video}, the proposed EMVPC dataset presents a deeper understanding about topical emotion, paragraph logic, relation inference and narrative description. Figure~\ref{fig:comparison-visual-fig1} shows two visualized examples of the proposed dataset.

Generally, EMVPC has the following three unique characteristics. First, EMVPC is paired with both English and Chinese paragraph captions, which can boost the development of vision-language study from monolingual to multilingual. Second, EMVPC contains abundant and comprehensive video contents, including 53 emotion categories and 376 realistic scenes. Third, the logic words in EMVPC enrich the lexical expression and support the generation of a paragraph with logical structure. Extensive experiments are conducted on two existing video paragraph captioning datasets (Charades Captions~\cite{wang2018video} and ActivityNet Captions~\cite{krishna2017dense}) as well as the proposed EMVPC. This work has evaluated 13 state-of-the-art models in terms of 15 popular metrics on these three datasets and performed comprehensive qualitative and quantitative analyses.

The main contributions of this work are summarized in the following three aspects.
\begin{itemize}
\item First, a full-fledge emotion and logic oriented video paragraph captioning dataset is constructed. The dataset with English-Chinese corpus contains 10,291 videos, 167.8 hours, 53 emotion classes and 376 realistic scenes, in which each emotion class is covered with several scenes. Moreover, the paragraph caption of each video is accompanied by emotion words and logic words to generate more logical and emotional expressions.

\item Second, a comprehensive video paragraph captioning research has been carried out that considers both emotions and logic in the narration. Existing state-of-the-art models are also evaluated in terms of 15 popular metrics on three challenging datasets (Charades Captions, ActivityNet Captions and the proposed EMVPC). An open-source benchmark library will be created to share the predicted results and the pre-trained models for all the competing methods with the research community.

\item Third, extensive experiments are carried out and the results are comprehensively analyzed in terms of qualitative and quantitative comparisons. Four important issues including efficiency, topicality, logicality and metric are investigated to provide the guidance for future research. This full-fledged dataset and associated study are pioneering attempts to promote paradigm-shifting advances in video paragraph captioning.
\end{itemize}

The rest of this paper is organized as follows. The related works in video paragraph captioning, video captioning datasets and video emotion recognition datasets are introduced in~\secref{sect:relwk}. The details of the EMVPC dataset including emotion and logic, data collection and dataset analysis are presented in~\secref{sect:emvpc}. The comprehensive quantitative and qualitative analyses of experimental results are described in~\secref{sect:expmt}. Discussion of important issues and future research directions are provided in~\secref{sect:discs}. Finally, we conclude this paper in~\secref{sect:conln}.

\section{Related Work}
\label{sect:relwk}

\subsection{Video Paragraph Captioning}
\label{sect:relwk:vpc}

Video paragraph captioning is the task of automatically describing video contents with multiple sentences and has achieved rapid development in the fields of computer vision and natural language processing. Different from the video captioning task that aims to output a simple and short sentence~\cite{yao2015describing,wang2018video,pan2020spatio,zhang2020object,zhang2021open}, the task of video paragraph captioning is more challenging for its multiple-sentence generation based on a long video with several events, and it can be generally divided into three categories. First, most popular methods~\cite{zhou2018end,wang2018bidirectional,MDVC_Iashin_2020,BMT_Iashin_2020} adopted a two-stage strategy that divided the event segments separately and described each segment with a sentence. These models concatenated individual sentences to obtain the final paragraph description. In~\cite{zhou2018end}, Zhou \textit{et al.} designed an end-to-end transformer model consisting of a proposal decoder and a captioning decoder to establish video event segments and restricted the attention of encoded representations. Wang \textit{et al.}~\cite{wang2018bidirectional} designed an attention fusion module and a context gating mechanism to effectively deal with event proposals and extracted video features. Iashin \textit{et al.}~\cite{BMT_Iashin_2020} developed a bi-modal transformer to construct a transformer framework to generate dense captions with audio, visual and textual inputs. Though each generated sentence is highly related to its event proposal, the final paragraph is not coherent or diverse after concatenating multiple sentences. Second, several recurrent-based models~\cite{xiong2018move,gella2018dataset,lei2020mart} leveraged the last hidden state of the decoder as the guidance of the next sentence prediction. Xiong \textit{et al.}~\cite{xiong2018move} proposed a recurrent network to jointly select a sequence of characteristic video clips and generated coherent sentences. Lei \textit{et al.}~\cite{lei2020mart} presented a novel approach called memory-augmented recurrent transformer to form a generalized memory state to encourage the next sentence production. These recurrent methods have shown good performance in processing multiple sentences, but they were still not able to obtain desirable results because the prior hidden state was only achieved at the last word or sentence level and the information degradation problem existed in the temporal dimension. Third, the one-stage model~\cite{song2021towards} is emerging and achieves superior performance. Song \textit{et al.}~\cite{song2021towards} firstly proposed a framework which abandoned the event detection step and directly outputted paragraph captioning. Despite of its diverse and coherent descriptions, this model is not capable of detecting precise event proposals and thus may produce captions that do not match the theme of a given video. Recently, several captioning formats are rapidly developing, such as emotion expressions~\cite{mathews2016senticap,wang2021emotion} and stylized representations~\cite{gan2017stylenet,chen2018factual}. In this work, an emotion and logic driven video paragraph captioning dataset is proposed to boost the fluency, logicality and abundance of the generated paragraph captions.

\subsection{Video Captioning Dataset}
\label{sect:relwk:vcd}

In order to advance the research in video captioning, numerous datasets covering different domains have been proposed, such as social media~\cite{gella2018a}, movies~\cite{rohrbach2015a,rohrbach2017movie}, cooking~\cite{da2013a,zhou2018towards,regneri2013grounding,rohrbach2014coherent} and human activities~\cite{chen2011collecting,sigurdsson2016hollywood,xu2016msrvtt,krishna2017dense,li2016tgif}. Generally, existing datasets for video captioning can be divided into single-sentence and multi-sentence counterparts. The characteristics of existing video captioning datasets~\cite{aafaq2019video} are summarized in Table~\ref{tab:comparison-datasets}. First, most datasets provide single sentence description~\cite{da2013a,wang2019vatex,chen2011collecting,xu2016msrvtt} to convey the information of the entire video clip. Chen~\textit{et al.}~\cite{chen2011collecting} proposed the MSVD dataset which consists of 1,970 video clips with simple and short sentences. Xu~\textit{et al.}~\cite{xu2016msrvtt} built a large-scale dataset named MSR-VTT, providing 10k web video clips with one annotated sentence for each video. Wang~\textit{et al.}~\cite{wang2019vatex} presented the VATEX dataset with 10 English and 10 Chinese captions per video. However, videos generally contain multiple activities or events, and a single sentence is typically not able to convey rich semantic information of the video. Second, multi-sentence datasets~\cite{regneri2013grounding,rohrbach2014coherent,sigurdsson2016hollywood,zeng2016generation,krishna2017dense,gella2018a,zhou2018towards} are introduced for a richer textual description. Regneri~\textit{et al.}~\cite{regneri2013grounding} extended the existing cooking corpus~\cite{rohrbach2012script} to multiple textual descriptions. Rohrbach~\textit{et al.}~\cite{rohrbach2014coherent} proposed the TSCoS-MultiLevel dataset which provides detailed description with multiple sentences for each video. Zhou~\textit{et al.}~\cite{zhou2018towards} constructed the YouCook-II dataset to elaborately record the cooking procedure in a video. Sigurdsson~\textit{et al.}~\cite{sigurdsson2016hollywood} proposed the Charades dataset in which caption is annotated with several sentences to depict human activities in the room. Nevertheless, these multi-sentence datasets are limited to specific domains (\textit{e.g.}, cooking and indoor). For open-domain datasets, VTW~\cite{zeng2016generation} is an ``in a wild'' video-language dataset that provides video titles and non-visual information in the description, but its primary goal is to generate video titles. Krishna~\textit{et al.}~\cite{krishna2017dense} introduced the ActivityNet Captions dataset for the dense-captioning task, but the narration of each video is not coherent because of the overlap between events. Gella~\textit{et al.}~\cite{gella2018a} proposed the VideoStory dataset that aims to tell a coherent story rather than a simple combination of multiple captions. Though multi-sentence datasets can record video contents with several captions, they either suffer from scene limitations of specific domains, or lack of informative cues (\textit{e.g.}, semantic coherence and rich emotions) in open domains. In order to tackle these problems, the EMVPC dataset is constructed for video paragraph captioning, in which the selected videos are from wide domains to cover a variety of human activities and emotions, the annotated paragraph captions are emotional, logical, coherent, elaborative and multilingual.

\begin{table}[htbp]
  \caption{Comparison of existing video description datasets. MLingual: multilingual dataset; multi-sent.: multiple-sentence description; len(hrs): duration of total videos (hours). - indicates that the data is not reported. \checkmark indicates datasets with multiple sentences, which can be used to generate multiple-sentence descriptions.}\vspace{-8pt}
  \label{tab:comparison-datasets}
  \centering
  \renewcommand{\arraystretch}{1.0}
  \renewcommand{\tabcolsep}{1pt}
  \newsavebox{\tableboxTabVS}
  \begin{lrbox}{\tableboxTabVS}
  \begin{tabular}{l|c|c|c|l|l}
  \hline
  \hline
    Dataset                                                                  & multi-sent.         & MLingual      &Domain         & videos:clips    & len(hrs)  \\
    \hline
    \hline
    YouCook~\cite{da2013a}                                     & -                       & -                  & cooking        & 88:-                 & 2.3       \\
    YouCook-II~\cite{zhou2018towards}                     & \checkmark       & -                  & cooking        & 2k:15.4k          & 176         \\
    TACoS~\cite{regneri2013grounding}                     & \checkmark       & -                  & cooking        & 127:3.5k          & 15.9       \\
    TACoS-M~\cite{rohrbach2014coherent}       & \checkmark       & -                  &cooking         & 185:25k           & 27.1        \\  
    MPII-MD~\cite{rohrbach2015a}                            & -    & -                  & movie           & 94:68k             & 73.6        \\
    M-VAD~\cite{torabi2015using}                             & -    & -                  & movie           & 92:49k             & 84.6         \\
    LSMDC~\cite{rohrbach2017movie}                       &  -    & -                  & movie           & 200:128k         & 150        \\
    Charades~\cite{sigurdsson2016hollywood}            & \checkmark       & -                  & indoor          & 10k:10k            & 82.01       \\
    VideoStory~\cite{gella2018a}                                & paragraph          & -                 & social media  & 20k:123k          & 396        \\
    VTW~\cite{zeng2016generation}                           & \checkmark       & -                 & open             & 18k:-                &  213.2       \\
    TGIF~\cite{li2016tgif}                                           & -                       & -                 & open             &     -:100k           &  86.1      \\
    VATEX~\cite{wang2019vatex}                             & -                       & \checkmark  & open            & 41.3k:41.3k       &  -       \\
    MSVD~\cite{chen2011collecting}                          & -                       & \checkmark  & open            & 2k:2k                & 5.3        \\
    MSR-VTT~\cite{xu2016msrvtt}                            & -                       & -                  & open            & 7k:10k              & 41.2        \\
    ANet-Cap~\cite{krishna2017dense}                  & \checkmark        & -                  & open            & 20k:100k           & 849.0       \\
    \hline
   EMVPC (Ours)                                                       & paragraph         & \checkmark   & open            & 10k:  31.7k                & 167.8   \\
   \hline
   \hline
   \end{tabular}
   \end{lrbox}
   \scalebox{0.95}{\usebox{\tableboxTabVS}}
\end{table}

\subsection{Video Emotion Recognition Dataset}
\label{sect:relwk:vdrd}

The psychological theory believes that emotions can be distinguished and divided into static categories on the basis of facial expressions, physiological measurements, behaviors and external causes~\cite{dolan2002emotion}. On one hand, according to the discrete emotion theory, all humans have a set of basic emotions that are innately and cross-culturally recognizable~\cite{Colombetti2009from}. Ekman~\cite{ekman1992argument} concluded six pan-cultural basic emotions, including anger, disgust, fear, happiness, sadness and surprise. Plutchik~\cite{plutchik2001the} diagramed a wheel of eight basic emotions, where anticipation and trust are added on the top of Ekman's six emotions. It is a conventional wisdom in psychology that humans are more likely to experience simultaneous emotions at a time while encountering external events, rather than just one certain basic emotion. The blend of emotions at three different levels (\textit{i.e.}, primary, secondary and tertiary dyads) are achieved through the pairwise combination on Plutchik's wheel of emotions~\cite{athar2011a}. For example, joy and trust lead to love, fear and sadness lead to despair, anticipation and fear lead to anxiety, etc. In this work, Plutchik's emotion theory is employed as the basis of static emotion categories, including basic and blended emotions. In addition, 9 empirical emotions in daily life are added, such as pain, excitement, embarrassment, etc.

On the other hand, as an essential and challenging direction in visual understanding, video emotion recognition has attracted great interest of researchers in recent years. The early works~\cite{kang2003affective,wang2006affective,irie2010affective} on emotion mainly focused on movies, which are limited to video scales and emotion types. Kang~\textit{et al.}~\cite{kang2003affective} collected various shots from six movies, in which three emotion categories were manually annotated. Wang~\textit{et al.}~\cite{wang2006affective} extracted 2,040 scenes from 36 popular Hollywood films as training data, and these scenes were labeled with 7 emotion categories. Irie~\textit{et al.}~\cite{irie2010affective} translated 24 movies into 206 scenes and labeled 8 emotion categories. Recent works on video emotion recognition introduced user-generated videos from sharing-websites, but the data scale and identifiable emotion categories were not significantly expanded, due to the difficulty of collecting and labeling. Jiang~\textit{et al.}~\cite{jiang2014predicting} collected videos from the Web and adopted 8 emotions based on the Plutchik's wheel. Xu~\textit{et al.}~\cite{xu2018heterogeneous} proposed two emotion-centric video datasets, YF-E6 and VideoStory-P14. YF-E6 was collected from social video-sharing websites through using 6 basic emotions as the keywords. VideoStory-P14 contains 626 videos belonging to 14 emotion classes, which was derived from the VideoStory dataset. In this work, the proposed EMVPC dataset is a large-scale emotion and logic driven dataset, containing over 10,000 videos and 53 emotion categories. As shown in Table~\ref{tab:comparison-emotions}, EMVPC is compared with existing video emotion recognition and emotion description datasets. Obviously, EMVPC is the largest benchmark in terms of video-emotion corpora.

\begin{table}[htbp]
  \caption{Comparison of existing video emotion recognition datasets and emotion description datasets. $\ddag$ denotes emotion description datasets, and - indicates that the item is not reported.
}
  \label{tab:comparison-emotions}
  \vspace{-8pt}
  \centering
  \renewcommand{\arraystretch}{1.0}
  \renewcommand{\tabcolsep}{1pt}
  \newsavebox{\tableboxTabVED}
  \begin{lrbox}{\tableboxTabVED}
  \begin{tabular}{l|c|c|c}
  \hline
  \hline
    Dataset                                                       &Domain            & Num   & Emotion classes \\
    \hline
    \hline
    Shots-3~\cite{kang2003affective}                                                         & movie             & -                        & 3   \\
    Scenes-7~\cite{wang2006affective}                                                        & movie             & 2,040                 & 7   \\
    Movie-8~\cite{irie2010affective}                                                            & movie            & 206                    & 8   \\
    YouTube-8~\cite{jiang2014predicting}                            & open              & 1,101                 & 8   \\
    YouTube-24~\cite{jiang2014predicting,xu2018heterogeneous}   & open              & 1,101                 & 24   \\
    YF-E6~\cite{xu2018heterogeneous}                                          & open              & 1,637                 & 6   \\
    VideoStory-P14~\cite{xu2018heterogeneous}                            & social media   & 626                    & 14   \\
    FlickrStyle10K$^{\ddag}$~\cite{gan2017stylenet}     &   open     & 10K     &2\\
    EmVidCap$^{\ddag} $~\cite{wang2021emotion}              &  open     & 1,897   & 34  \\
   \hline
   EMVPC (Ours)                                                                            & open              & 10,291               & 53   \\
   \hline
   \hline
   \end{tabular}
   \end{lrbox}
   \scalebox{0.95}{\usebox{\tableboxTabVED}}
\end{table}

\section{The EMVPC Dataset}
\label{sect:emvpc}

As aforementioned, the proposed EMVPC dataset aims to provide a large-scale emotion and logic driven multilingual dataset to encourage the generation of emotional and logical video paragraph captions. The details of EMVPC are presented below.

\subsection{Emotion and Logic}
\label{sect:emotions}

\begin{figure}[htpb]
\begin{center}
\includegraphics[width=1\linewidth]{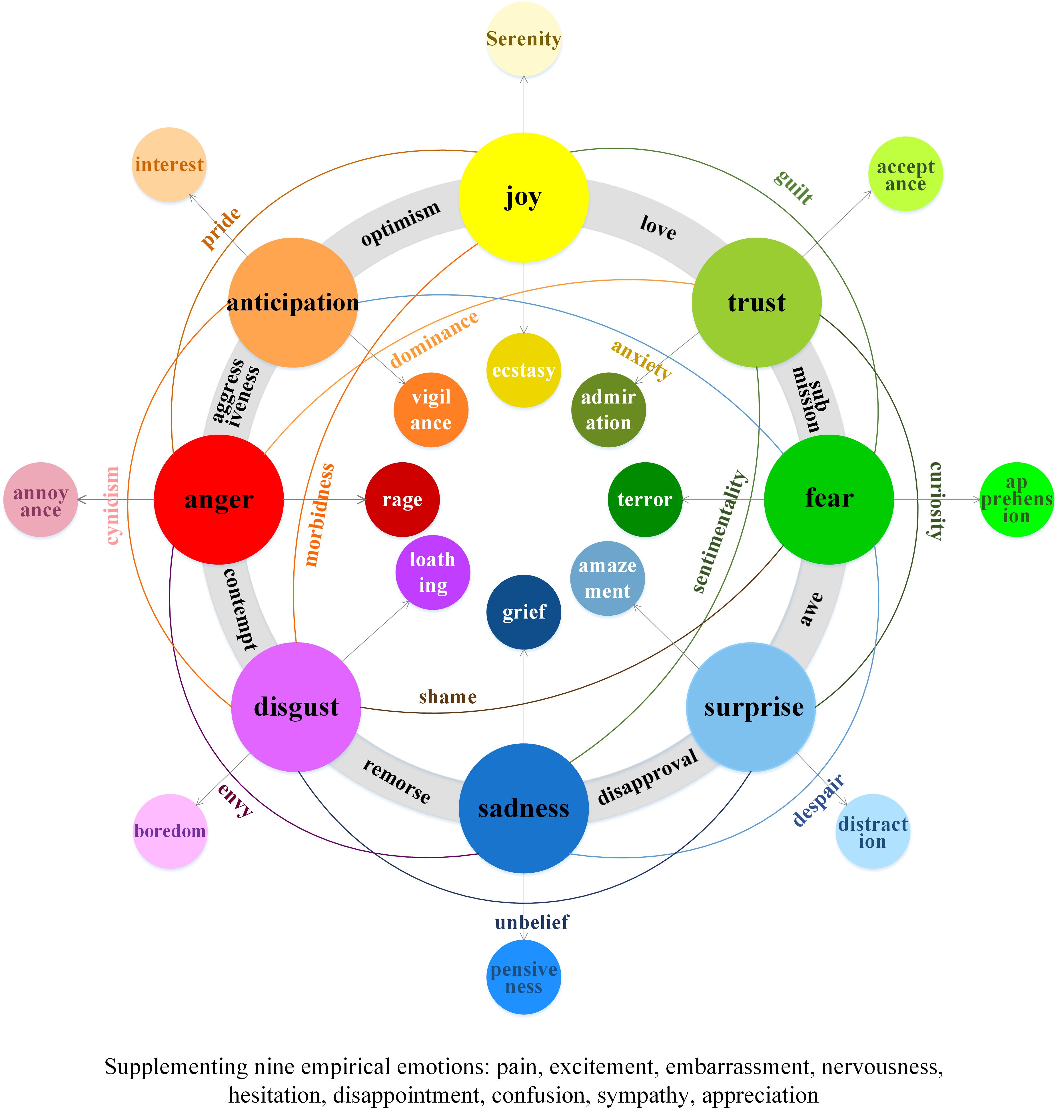}
\end{center}
\caption{The proposed EMVPC dataset includes 53 emotion categories, where 44 emotions are automatically generated according to the theory in~\cite{plutchik2001the} and another 9 emotions are manually added.}
\label{fig:emotion}
\end{figure}

Emotion is an essential element in user-generated videos, as well as movies and TV shows. In order to build a dataset that embodies rich-emotion videos, as shown in Fig.~\ref{fig:emotion}, 53 human emotions based on the psychological theories~\cite{plutchik2001the,athar2011a} are selected as the primary keywords for video collection. Because human emotional perception is subjective, different annotators may have different understandings of the emotions of a video. To avoid this cognitive conflict as much as possible, various realistic scenes are selected to support emotion identification in videos and are regarded as the secondary keywords for video collection. The whole process can be divided into two stages: emotion selection and scene collection.

In the first stage, according to Plutchik's wheel of emotions~\cite{plutchik2001the}, 8 basic emotions (\textit{i.e.}, joy, trust, fear, surprise, sadness, disgust, anger and anticipation) are chosen. Then, 8 stronger emotions (\textit{i.e.}, ecstasy, admiration, terror, amazement, grief, loathing, rage and vigilance) and 8 weaker ones (\textit{i.e.}, serenity, acceptance, apprehension, distraction, pensiveness, boredom, annoyance and interest) corresponding to the above 8 basic emotions are selected. Furthermore, 20 blended emotions at three different levels are employed, including 8 primary dyads (\textit{i.e.}, love, submission, awe, disapproval, remorse, contempt, aggressiveness and optimism), 7 secondary dyads (\textit{i.e.}, guilt, curiosity, despair, unbelief, envy, cynicism and pride) and 5 tertiary dyads (\textit{i.e.}, sentimentality, shame, morbidness, dominance and anxiety). As shown in Fig.~\ref{fig:emotion}, the primary dyad is a mixture of two neighboring basic emotions (\textit{e.g.}, fear and surprise lead to awe), the secondary dyad is a mixture of two basic emotions separated by one emotion (\textit{e.g.}, anger and joy lead to pride) and the tertiary dyad is a mixture of two basic emotions separated by two emotions (\textit{e.g.}, fear and disgust lead to shame) according to the rules in~\cite{athar2011a, turner2000on}. Note that ``hope'' from the secondary dyad, ``delight'', ``outrage'' and ``pessimism'' from the tertiary dyad are abandoned, because they are similar to the selected emotions. Moreover, 9 emotions which usually happen in daily life (\textit{i.e.}, pain, excitement, embarrassment, nervousness, hesitation, disappointment, confusion, sympathy and appreciation) are added. In total, 53 emotions are used for the construction of EMVPC. In the second stage, a number of candidate scenes are firstly listed for each emotion to choose, and a number of professional volunteers are employed to update the scene list for each emotion. During the decision of scene list, it is required to ensure that each emotion relates to 6-9 scenes. Note that scenes are added or deleted according to the actual situation in the collection procedure, and the final number of scenes is 367 after completing video collection.

Logic plays a key role in forming the structure of human language, boosting the contextual relevance and fluent expression which are important for understanding. In a paragraph, the logical connectors~\cite{arapoff1968the} can reflect the logical relationships between sentences. According to the logical connectors in~\cite{arapoff1968the,quirk1982a,celce1983the,narita2004connector,bolton2002a}, 17 customized logical relationships are summarized in Table~\ref{tab:logic}, and each logical relationship is given a detailed definition with characteristics, related logical connectors and exemplary sentences to assist annotation. Note that there are inclusive or supplementary relations between some logical relationships, such as adversative relation and concession relation, ordinal relation and time relation, so several connectors belonging to different logical relations may overlap, but annotation is not affected. As shown in Table~\ref{tab:logic}, the logical connectors corresponding to each logical relationship are presented, including adversative relation, concession relation, contrast relation, comparison relation, parallel relation, ordinal relation, progressive relation, illustration relation, causal relation, purpose relation, conditional relation, negative relation, interpretation relation, alternative relation, summary relation, time relation and reporting relation. Specifically, in order to enhance the logic of paragraph description in EMVPC, annotators are required to describe paragraph with at least one logical connector based on the designed logical relationships.

\begin{table}[htbp]
  \caption{Logical relationships and examples of logical connectors.}
  \label{tab:logic}
  \vspace{-8pt}
  \centering
  \renewcommand{\arraystretch}{1.0}
  \renewcommand{\tabcolsep}{1pt}
  \newsavebox{\tableboxTabLOG}
  \begin{lrbox}{\tableboxTabLOG}
  \begin{tabular}{c|l|l}
  \hline
  \hline
    \#  & Logical relationships  & Examples of logical connectors  \\
  \hline
  \hline

          1 & adversative        & but, however, while, despite, yet, nevertheless\\
          2 & concession        & although, though, even though, even if\\
          3 & contrast             & in contrast, on the contrary, rather than, far from\\
          4 & comparison        & bigger than, more than, as...as \\
          5 & parallel               & and, as well as, likewise, equally, neither...nor \\
          6 & ordinal               & after, before, when, then, first-second-last of all \\
          7 & progressive        & even, indeed, in addition, furthermore, besides \\
          8 & illustration          & such as, for example, for instance, namely\\
          9 & causal                & because, hence, due to, as a result, so that, therefore \\
          10 & purpose           & in order to, so that, for, so as to \\
          11 & condition         & if, unless, once, depending on, as long as, whatever\\
          12 & negative           & not, never, none, neither, anything...but, too...to\\
          13 & interpretation    & that is, which means, to explain \\
          14 & alternative         & either...or, would rather...than, prefer...to, instead of\\
          15 & summary          & in all, in brief, altogether, in conclusion, to summarize\\
          16 & time                 & at the same time, immediately, afterwards, later\\
          17 & reporting          & quote, report\\
\hline
\hline
\end{tabular}
\end{lrbox}
\scalebox{0.92}{\usebox{\tableboxTabLOG}}
\end{table}

\subsection{Data Collection}
\label{sect:collection}

The videos in EMVPC are collected from two sources: (1) major video sharing websites on the Internet (\textit{e.g.}, YouTube\footnote{https://www.youtube.com/}, YOUKU\footnote{https://www.youku.com/}), and (2) popular TV shows and movies. In order to make it easier to distinguish the emotion in videos, emotion and scene are considered at the same time to collect videos as candidates. The collectors can choose any Internet videos, popular TV shows and movies except those involving sexism, racial discrimination, political sensitivity, pornographic and violence. The duration of a video is required to be controlled between 30 seconds and 5 minutes. Each video is divided into several segments which are then translated into corresponding natural language annotations with emotions and logic. The entire data collection and paragraph annotation take around 21 months.

To ensure the high quality of EMVPC, annotators are native Chinese who are graduates in English major or have professional English certificates and skills. The annotation process is divided into three steps. First, annotators are required to watch the video. Next, the video is semantically split into 2 to 9 segments, and each segment is described by sentence in English and Chinese versions. Finally, these sentences are combined into a paragraph for further polishing. Especially, a paragraph is required to contain at least one emotion word for reflecting the topical emotion, and logical connection words should be embodied to highlight the logic of the paragraph.

To further improve dataset quality, a strict verification process is conducted first. Each video is reviewed by 2 to 4 volunteers and scored according to the relevance between video content and emotion label. The score is divided into five grades, concretely, identical (corresponding to 10 points), related (corresponding to 9, 8, 7 points), almost (corresponding to 6, 5, 4 points), bare (corresponding to 3, 2, 1 points) and unrelated (corresponding to 0 point). When the average score of a video is less than 7 points, or there are more than 3 points between the highest and the lowest scores, the video will be reviewed and discussed again. Second, a comprehensive review process is implemented for paragraph verification, which contains 3 evaluation criteria (\textit{i.e.}, the rationality of video segmentation, the correctness of emotion and logic, and the accuracy and richness of paragraph expression). Moreover, these criteria are specifically divided into the following 11 aspects: (1) whether each segment's starting and ending times are reasonable, (2) whether the caption and the corresponding segment are semantically consistent, (3) whether the paragraph narrative contains the corresponding emotion, (4) whether the logic in the paragraph narrative is reasonable, (5) the correctness of emotion words, (6) the correctness of logic words, (7) the tenses in English, (8) whether the paragraph is too complex or too colloquial, (9) objective expression (accuracy, fluency, logic), (10) whether the detailed description is rich and delicate and (11) whether emotional vocabularies are diverse. The quality of each annotated caption is measured based on scores summed from the above 11 aspects, and the higher score means the better annotation quality.

\subsection{Dataset Analysis}
\label{data analysis}

The EMVPC dataset contains 10,291 videos, and each video is annotated with one English and one Chinese paragraph captions. All videos are emotion related and the video number of each emotion is shown in Table~\ref{tab:video nums}. In general, some emotions that people easily expose tend to associate with more videos, such as joy, anger, fear, etc. On the contrary, some emotions that are hard to detect in videos relate to fewer videos, such as sympathy, appreciation, envy, etc. Specifically, more than 50$\%$ and nearly 70$\%$ of emotions relate to more than 100 and 50 videos, respectively. Each emotion relates to an average of 7.1 scenes, but there are big differences in the number of scenes covered by different emotions. For common emotions (\textit{e.g.}, joy, fear), they relate to 21 and 15 scenes, respectively. While for some emotions that are difficult to collect, such as unbelief, envy and cynicism, they only relate to one scene. Overall, more than 50$\%$ of emotions relate to more than 7 scenes.

\begin{table}[htbp]
  \caption{Statistics of total video number of each emotion.}\vspace{-8pt}
  \label{tab:video nums}
  \centering
  \renewcommand{\arraystretch}{1.0}
  \renewcommand{\tabcolsep}{1pt}
  \newsavebox{\tableboxTabVNUM}
  \begin{lrbox}{\tableboxTabVNUM}
  \begin{tabular}{l|l|c|l|l|c|l|l|c}
  \hline
  \hline
    \# & emotion & num & \# & emotion & num & \# & emotion & num \\
    \hline
    \hline
    01 & joy          & 1,114    & 21 & grief      & 134     & 41 & cynicism       & 46   \\
    02 & trust        &  166   & 22 & loathing     & 47     & 42 & sentimentality & 15   \\
    03 & fear         & 750    & 23 & rage         & 299     & 43 & morbidness     & 12   \\
    04 & surprise     & 643    & 24 & vigilance    & 145     & 44 & dominance      & 30   \\
    05 & sadness      & 677    & 25 & love         & 753     & 45 & pain           & 180   \\
    06 & disgust      & 393    & 26 & submission   & 95     & 46 & excitement     & 152   \\
    07 & anger        & 841    & 27 & awe          & 76     & 47 & embarrassment  & 70   \\
    08 & anticipation & 183    & 28 & disapproval  & 139     & 48 & nervousness    & 88   \\
    09 & serenity     & 101    & 29 & remorse      & 30     & 49 & hesitation     & 26    \\
    10 & acceptance   & 84    & 30 & contempt     & 161     & 50 & disappointment & 20    \\
    11 & apprehension & 174    & 31 & aggressiveness& 345    & 51 & confusion      & 46    \\
    12 & distraction  & 95    & 32 & optimism      & 138    & 52 & sympathy       & 7    \\
    13 & pensiveness  & 159    & 33 & anxiety       & 380    & 53 & appreciation   & 8     \\
    14 & boredom      & 409    & 34 & pride         & 21    &  &      &      \\
    15 & annoyance    & 161    & 35 & despair       & 96    &  &      &      \\
    16 & interest     & 29    & 36 & shame         & 36    &  &      &     \\
    17 & ecstasy      & 104    & 37 & guilt         & 71    &  &      &      \\
    18 & admiration   & 18    & 38 & curiosity     & 126    &  &      &      \\
    19 & terror       & 108    & 39 & unbelief      & 63    &  &      &      \\
    20 & amazement    & 218    & 40 & envy          & 9    &  &      &     \\
    \hline
    \hline
    \end{tabular}
    \end{lrbox}
    \scalebox{0.95}{\usebox{\tableboxTabVNUM}}
\end{table}

\begin{figure}[htbp]
\begin{center}
\includegraphics[width=1\linewidth]{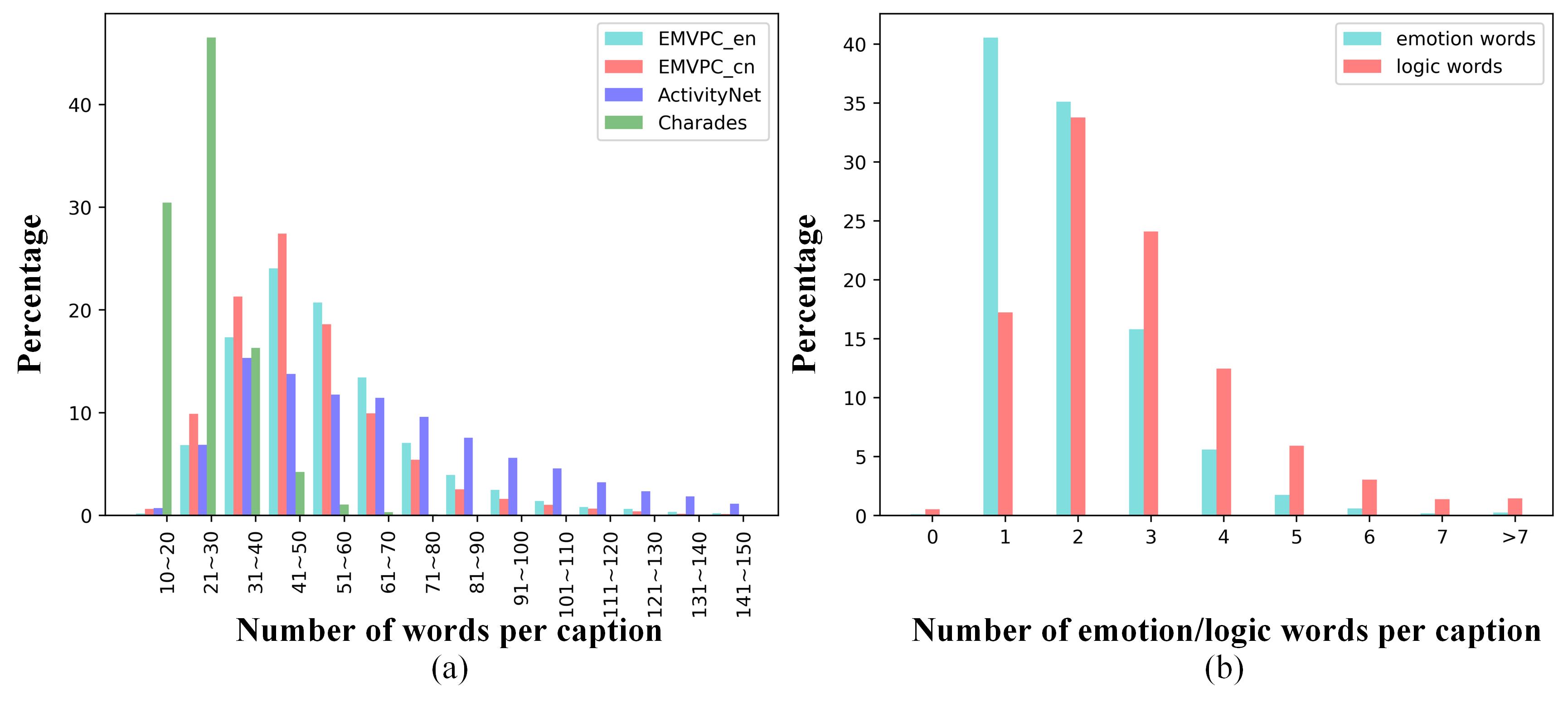}
\end{center}
\caption{Statistic histogram distribution of the proposed EMVPC. (a) Distribution of the number of words per caption. (b) Distribution of the number of emotion and logic words per caption.}
\label{fig:histogram}
\end{figure}

Figure~\ref{fig:histogram}(a) compares the proposed dataset with the Charades Captions dataset~\cite{wang2018video} and the ActivityNet Captions dataset~\cite{krishna2017dense} on the number of words per caption. It can be observed that the captions of EMVPC-en and EMVPC-cn are longer than Charades Captions and shorter than ActivityNet Captions. The short videos in Charades  Captions usually generate brief descriptions and the long videos in ActivityNet Captions tend to produce dense captions according to their segmented clips. Different from the rough or redundant descriptions in previous datasets, the number of words in each sentence of EMVPC is controlled from 8 to 35, and if it is more than 35 words, annotators are recommended to divide segments and translate them again. Therefore, a concise yet comprehensive paragraph caption is provided for each video, and the paragraph length is mainly in the range of 21 to 70 words. Figure~\ref{fig:histogram}(b) reports the number of emotion and logic words per caption in EMVPC. It can be seen that each caption usually contains 1-4 emotion and logic words, indicating the rich emotional and logical information in the caption.

\subsection{Data Split}

Experiments are conducted on the Charades Captions dataset~\cite{wang2018video}, the ActivityNet Captions dataset~\cite{krishna2017dense} and the proposed EMVPC dataset. Charades Captions is consisted of 6,963 training videos, 500 validation videos and 1,760 testing videos, and each video is annotated with multiple sentences. ActivityNet Captions contains 10,009 training videos, 4,917 validation videos and 5,044 testing videos, each training video has one annotated paragraph and each validation video has two annotated paragraphs; following the previous work~\cite{song2021towards}, the validation set is split into two components: 2,460 \textit{ae-val} videos and 2,457 \textit{ae-test} videos. The proposed EMVPC has 8,232 training videos and 2,059 testing videos, and each video contains one English paragraph and one Chinese paragraph.
\section{Benchmark Experiment}
\label{sect:expmt}

\subsection{Experimental Setting}

\subsubsection{Evaluation Metric}

Comprehensive experiments are conducted to evaluate the quantitative performance on the proposed EMVPC in terms of the following metrics, including BLEU~\cite{papineni2002bleu}, METEOR~\cite{banerjee2005meteor}, ROUGE\_L~\cite{lin2004rouge}, CIDEr~\cite{vedantam2015cider} and SPICE~\cite{anderson2016spice}. The goal of these metrics is to provide the measurement of semantic correlations between predicted sentences and annotated references. Specifically, BLEU~\cite{papineni2002bleu} primarily measures the n-gram overlapping between candidates and annotations, and more number of matches usually means the better translation quality of predictions. METEOR~\cite{banerjee2005meteor} explores the unigram matching between machine-generated translation and human-generated translation, and precision, recall and a metric of fragmentation are combined to compute METEOR. ROUGE\_L~\cite{lin2004rouge} computes the longest common subsequence of candidates and references, and precision, recall and a weighting factor are merged to obtain the ROUGE\_L score. CIDEr~\cite{vedantam2015cider} is a customized measurement for visual captioning evaluation, which calculates the cosine similarities between predictions and references to measure the consistency of annotations. SPICE~\cite{anderson2016spice} utilizes the scene graph to obtain the semantical embeddings of objects, attributes and relationships, then the score that corresponds to human judgment standard is calculated. In addition, a language generation based metric called BERTScore~\cite{zhang2019bertscore} is rapidly emerging in the field of machine translation, which extracts the token embeddings of each candidate and reference sentence using the pre-trained BERT~\cite{devlin2018bert} network, and then computes their token similarity for obtaining the recall ($R_{bert}$), precision ($P_{bert}$) and F-score ($F_{bert}$), respectively. BERTScore is able to measure the sentence-level semantic similarity between the generated paragraph and the reference. The higher the BERTScore is, the better logic the generated paragraph contains. Moreover, in order to evaluate emotion accuracy of generated sentences, ACC$_{sw}$, ACC$_{c}$, BFS and CFS are proposed in~\cite{wang2021emotion}, where ACC$_{sw}$ and ACC$_{c}$ calculate the accuracy of emotion words and emotion sentences, respectively, and BFS as well as CFS give a comprehensive evaluation of generated sentences in terms of facts and emotions.

\begin{figure*}[htbp]
\begin{center}
\begin{tabular}{c}
\includegraphics[width=0.9\linewidth]{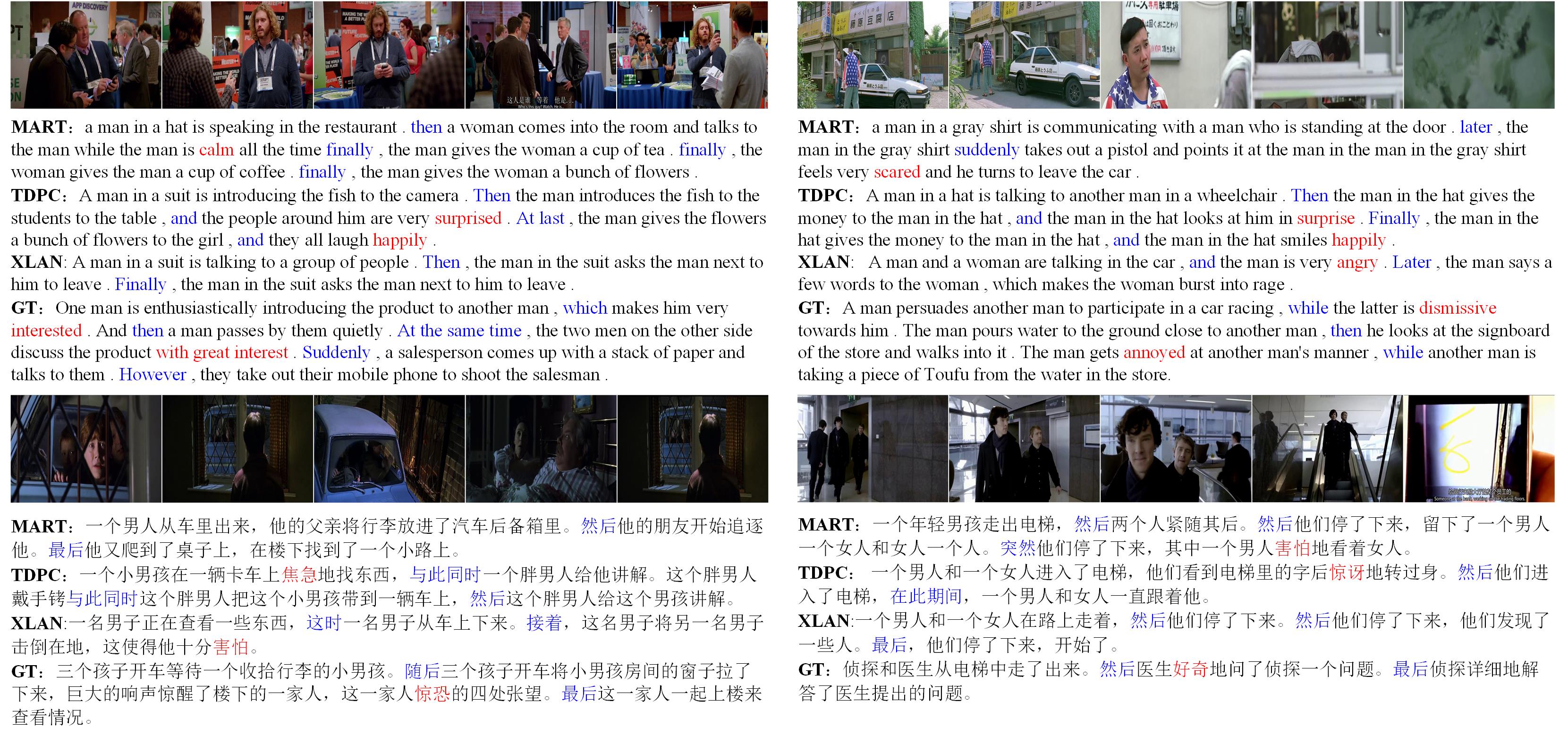}
\end{tabular}
\end{center}
   \caption{Visualized comparisons of MART, TDPC, XLAN and ground-truth, where videos in the upper part and lower part are the results of English version and Chinese version of EMVPC, respectively. The emotion words and the logic words are highlighted in \textcolor[rgb]{1.00,0.00,0.00}{red} and \textcolor[rgb]{0.00,0.00,1.00}{blue}.}
\label{fig:comparison-visual}
\end{figure*}

\subsubsection{Competitor}

In the video paragraph captioning task, 13 state-of-the-art models are selected for performance comparison, including 4 image captioning methods (Transformer~\cite{sharma2018conceptual}, M2NET~\cite{cornia2020meshed}, XTransformer~\cite{pan2020x}, XLAN~\cite{pan2020x}), 2 video captioning approaches (MP-LSTM~\cite{venugopalan2014translating}, TA~\cite{yao2015describing}), 1 vision-language pre-trained model (UniVL~\cite{luo2020univl}), 3 dense video captioning methods (VTransformer~\cite{zhou2018end}, MDVC~\cite{MDVC_Iashin_2020}, BMT~\cite{BMT_Iashin_2020}) and 3 video paragraph captioning methods (AdvInf~\cite{park2019adversarial}, MART~\cite{lei2020mart}, TDPC~\cite{song2021towards}). These models are chosen following two principles: (1) representative and (2) source codes and results.

\subsubsection{Benchmark Protocol}

The competing methods are evaluated on three video paragraph captioning datasets including Charades Captions~\cite{wang2018video}, ActivityNet Captions~\cite{krishna2017dense} and the proposed EMVPC. To the best of our knowledge, the largest and most extensive video paragraph captioning benchmark is established in this work. To compare fairly, the released codes are directly trained and inferred with official settings.

\subsection{Qualitative Comparison}

Four visualized groups of the three methods MART~\cite{lei2020mart}, TDPC~\cite{song2021towards} and XLAN~\cite{pan2020x} are presented in Fig.~\ref{fig:comparison-visual}. Generally, benefitting from rich emotional and logical representations of the proposed EMVPC, the generated sentences are more abundant, coherent and narrative than the generated paragraph captions based on traditional captioning datasets. First, there are more emotion expressions in the generated sentences. For example, in the top-left video, TDPC vividly depicts human emotions with \textit{``surprise"} and \textit{``happily"}. Second, the cohesion among the generated sentences becomes smoother, and the logical relationship is clearer. For example, in the top-left video, all of these three methods generate ordinal logical connectors like \textit{``then-finally"} and \textit{``then-at last"}. In the top-right video, besides ordinal connectors, time connectors \textit{``suddenly"} and \textit{``later"} are also generated. 

However, several apparent problems still exist in the produced sentences. First, the captioning models cannot identify the appropriate emotions that occur in untrimmed video clips. For example, in the top-left and bottom-right videos, the XLAN model~\cite{pan2020x} misses emotions. In the top-left video, the MART model~\cite{lei2020mart} misunderstands the thematic emotion \textit{``interest"}, describing the emotion with \textit{``calm"} instead. Second, the produced paragraph is lack of reasonable and fluent description. For the MART model in the top-right video, the generation has grammatical mistakes and the logical associations among sentences are unreasonable, such as the description \textit{``points it at the man in the man in the gray shirt feels very scared and he turns to leave the car''}. Third, existing approaches restrict the diverse representations of the generated sentences. In the top-left video, the second and third sentences of XLAN exhibit the repetitive sentence \textit{``in the suit asks the man next to him to leave''}. Such situations damage the plentiful description of video paragraph, and there still exists huge room for exploring advanced schemes.

\begin{table*}[htbp]
  \caption{ Benchmark results of existing state-of-the-art approaches on Charades Captions and ActivityNet Captions. The bold font indicates the best performance. $\ddag$ denotes the model using segment annotations in both training and testing procedures. B@1, B@2, B@3, B@4, M, R, C, S represent the scores of BLEU-1, BLEU-2, BLEU-3, BLEU-4, METEOR, ROUGE\_L, CIDEr and SPICE, respectively.
  }\vspace{-8pt}
  \label{tab:comparision-anet-charades}
  \centering
  \renewcommand{\arraystretch}{1.0}
  \renewcommand{\tabcolsep}{2.4pt}
  \newsavebox{\tableboxTabRESCA}
  \begin{lrbox}{\tableboxTabRESCA}
  \begin{tabular}{ll||ll||cccccccc||ccc}
  \hline
  \hline
    &Methods& Year & Params & B@1  & B@2 & B@3 & B@4  & M & R & C  & S  & $P_{bert}$ & $R_{bert}$ & $F_{bert}$\\
    \hline
    \hline
    \multirow{10}{*}{\begin{sideways}\small{Charades Captions}\end{sideways}}
    &MP-LSTM~\cite{venugopalan2014translating} &NAACL2015 &185.8M &0.6173 &0.4215 &0.2791 &0.1826 &0.4082 &0.4259 &0.2220 &0.1427 &0.9010 &0.8889 &0.8948 \\
    &TA~\cite{yao2015describing} &ICCV2015 &247.2M &0.6228 &0.4174 &0.2831 &0.1855 &0.4129 &0.4285 &0.2344 &0.1484 &0.9009 &0.8893 &0.8950 \\
    &Transformer~\cite{sharma2018conceptual} &ACL2018 &226.2M &\textbf{0.6522} &\textbf{0.4513} &0.3124 &0.2096 &0.4353 &0.4452 &\textbf{0.2942} &\textbf{0.1737} &0.9057 &0.8925 &\textbf{0.8990 } \\
    &UniVL ~\cite{luo2020univl} & arxiv2020 &774.9M &0.6372 &0.4447& \textbf{0.3140} &\textbf{0.2156}& \textbf{0.4392} &\textbf{0.4534} &0.2867& 0.1719& \textbf{0.9067} &0.8909& 0.8986\\
    &M2NET~\cite{cornia2020meshed} & CVPR2020 &124.5M &0.5877 &0.3826 &0.2480 &0.1534 &0.4013& 0.4103 &0.2220& 0.1567& 0.9001& 0.8872& 0.8935  \\
    &MDVC~\cite{MDVC_Iashin_2020} & CVPRW2020 &244.3M &0.5973& 0.3841& 0.2519 &0.1577 &0.3984 &0.4161 &0.1933 &0.1432 &0.8977 &0.8885 &0.8930  \\
    &XTransformer~\cite{pan2020x} &CVPR2020&299.8M &0.6308& 0.4144& 0.2759 &0.1762 &0.4099 &0.4250 &0.2013 &0.1434 &0.8997 &0.8903 &0.8949 \\
    &XLAN~\cite{pan2020x} & CVPR2020 &511.5M& 0.6408 &0.4386& 0.2992& 0.1971& 0.4284 &0.4387& 0.2861 &0.1706 &0.9037 &\textbf{0.8935} &0.8985 \\
    &BMT~\cite{BMT_Iashin_2020}  & BMVC2020 &251.2M &0.6146 &0.3997& 0.2643& 0.1687& 0.4033 &0.4214 &0.2031& 0.1480 &0.8982 &0.8899 &0.8939\\
    &TDPC~\cite{song2021towards} & CVPR2021 &146.0M &0.6368 &0.4363 &0.3021 &0.2004 &0.4253 &0.4402& 0.2781 &0.1692 &0.9020 &0.8928 &0.8973  \\
    \hline
    \hline
    \multirow{13}{*}{\begin{sideways}\small{ActivityNet Captions}\end{sideways}}
    &MP-LSTM~\cite{venugopalan2014translating} &NAACL2015  & 422.2M &0.4319 &0.2420 &0.1489 &0.0965 &0.2909 &0.3013 &0.1256 &0.1166 &0.8774 &0.8646 &0.8708\\
    &TA~\cite{yao2015describing} &ICCV2015  & 483.6M &0.4450 &0.2579 &0.1621 &0.1061& 0.3073 &0.3132 &0.1643 &0.1261& 0.8800& 0.8658 &0.8727\\
    &Transformer~\cite{sharma2018conceptual} &ACL2018  &344.5M &0.4869 &0.3050 &0.1947 &0.1261 &0.3470 &\textbf{0.3459} &0.2570& 0.1647 &0.8888& 0.8701 &0.8793 \\
    &VTransformer$^{\ddag}$~\cite{zhou2018end} & CVPR2018 & 275.6M &0.4999 &0.2923 &0.1723 &0.1038 &0.3354 &0.3123 &0.2098 &0.1609 &0.8801 &0.8726 &0.8762 \\
    &AdvInf$^{\ddag}$~\cite{park2019adversarial} & CVPR2019 & 200.8M &0.4873 &0.2927 &0.1776 &0.1089 &0.3436 &0.3381 &0.2040 &0.1692 &0.8820 &\textbf{0.8784} &0.8801 \\
    &UniVL~\cite{luo2020univl} & arxiv2020 &774.9M  & 0.4803 &  0.2905 &  0.1832  & 0.1194  & 0.3285  & 0.3271  & 0.2255  & 0.1490  & 0.8816  & 0.8693  & 0.8753 \\
    &M2NET~\cite{cornia2020meshed} & CVPR2020 &154.7M &0.4852& 0.2776& 0.1603& 0.0949& 0.3185 &0.3110& 0.2020 &0.1480& 0.8800 &0.8721 &0.8759 \\
    &MDVC~\cite{MDVC_Iashin_2020} & CVPRW2020 &302.1M &0.5030 &0.2927 &0.1733 &0.1051 &0.3291 &0.3239 &0.2190 &0.1482 &0.8830 &0.8738 &0.8783 \\
    &XTransformer~\cite{pan2020x} & CVPR2020 &563.0M &0.4802 &0.2792 &0.1709 &0.1086 &0.3246 &0.3198 &0.2034 &0.1379 &0.8851 &0.8745 &0.8797 \\
    &XLAN~\cite{pan2020x} & CVPR2020  &308.0M &0.5065 &\textbf{0.3157}& \textbf{0.2034} &\textbf{0.1338} &0.3509 &0.3453 &0.2682 &0.1708 &0.8885 &0.8746 &0.8814\\
    &BMT~\cite{BMT_Iashin_2020}  & BMVC2020   &311.7M &0.5055 &0.2950 &0.1759 &0.1074 &0.3312 &0.3246 &0.2302 &0.1508 &0.8838 &0.8745 &0.8790\\
    &MART$^{\ddag}$~\cite{lei2020mart} & ACL2020   &110.8M &\textbf{0.5198} &0.3135 &0.1931 &0.1220 &\textbf{0.3544} &0.3449 &0.2601 &\textbf{0.1710} &0.8882 &0.8781 &0.8830\\
    &TDPC~\cite{song2021towards} & CVPR2021 &182.4M &0.5085 &0.3070 &0.1911 &0.1218 &0.3463& 0.3361 &\textbf{0.2737} &0.1666 &\textbf{0.8890}& 0.8780 &\textbf{0.8833} \\
    \hline
    \hline
    \end{tabular}
    \end{lrbox}
    \scalebox{1.0}{\usebox{\tableboxTabRESCA}}
\end{table*}

\begin{table*}[htbp]
  \caption{ Benchmark results of existing state-of-the-art approaches on the proposed EMVPC dataset. The bold font indicates the best performance. $\ddag$ denotes the model using segment annotations in both training and testing procedures, - means the statistic results are not available. B@1, B@2, B@3, B@4, M, R, C, S represent the scores of BLEU-1, BLEU-2, BLEU-3, BLEU-4, METEOR, ROUGE\_L, CIDEr and SPICE, respectively.
  }\vspace{-8pt}
  \label{tab:comparision}
  \centering
  \renewcommand{\arraystretch}{1.0}
  \renewcommand{\tabcolsep}{2.4pt}
  \newsavebox{\tableboxTabRESEM}
  \begin{lrbox}{\tableboxTabRESEM}
  \begin{tabular}{ll||ll||cccccccc||ccc}
  \hline
  \hline
    &Methods& Year & Params & B@1  & B@2 & B@3 & B@4  & M & R & C  & S  & $P_{bert}$ & $R_{bert}$ & $F_{bert}$\\
    \hline
    \hline
    \multirow{11}{*}{\begin{sideways}\small{English}\end{sideways}}
    &MP-LSTM~\cite{venugopalan2014translating} &NAACL2015 & 347.7M&0.2423 &0.1361 &0.0792 &0.0487 &0.3025 &0.2771 &0.0638 &0.1185 &0.8920 &0.8688 &0.8801\\
    &TA~\cite{yao2015describing} &ICCV2015  &  409.2M             &0.2641 &0.1513 &0.0907 &0.0574 &0.3161 &0.2880 &0.0768 &0.1306 &\textbf{0.8945} &0.8710 &0.8825\\
    &Transformer~\cite{sharma2018conceptual} &ACL2018  &307.3M    &0.3658 &0.2169 &0.1352 &0.0873 &\textbf{0.3421} &\textbf{0.3087} &0.1550 &0.1546 &0.8920 &0.8776 &0.8846 \\
    &UniVL~\cite{luo2020univl} & arxiv2020 &774.9M                &0.3517 &0.1883 &0.1090 &0.0663 &0.3072 &0.2834 &0.1465 &0.1506 &0.8817 &0.8733 &0.8774 \\
    &M2NET~\cite{cornia2020meshed} & CVPR2020 &147.0M             &0.3411 &0.1900 &0.1118 &0.0691 &0.3202 &0.2790 &0.1519 &0.1397 &0.8856 &0.8766 &0.8810 \\
    &MDVC~\cite{MDVC_Iashin_2020} & CVPRW2020 &295.0M             &0.3866 &0.2158 &0.1271 &0.0780 &0.3196 &0.2899 &0.1396 &0.1297 &0.8836 &0.8785 &0.8810 \\
    &XTransformer~\cite{pan2020x} & CVPR2020 &551.6M             &0.3490 &0.1920 &0.1116 &0.0689 &0.3093 &0.2827 &0.1132 &0.1188 &0.8861 &0.8764 &0.8811 \\
    &XLAN~\cite{pan2020x} & CVPR2020  &292.7M                     &0.3640 &0.2171 &\textbf{0.1362} &\textbf{0.0889} &0.3405 &0.3084 &0.1747 &\textbf{0.1572} &0.8915 &0.8786 &\textbf{0.8849} \\
    &BMT~\cite{BMT_Iashin_2020}  & BMVC2020   &304.6M             &\textbf{0.3899}&\textbf{0.2188} &0.1295 &0.0803 &0.3299 &0.2919 &0.1513 &0.1331 &0.8845 &0.8786 &0.8814 \\
    &MART$^{\ddag}$~\cite{lei2020mart} & ACL2020   &129.7M        &0.3865 &0.2063 &0.1168 &0.0690 &0.3273 &0.2820 &\textbf{0.2347} &0.1473 &0.8824 &0.8767 &0.8795 \\
    &TDPC~\cite{song2021towards} & CVPR2021 &178.9M               &0.3779 &0.2137 &0.1271 &0.0786 &0.3274 &0.2949 &0.1403 &0.1423 &0.8885 &\textbf{0.8792} &0.8837 \\
    \hline
    \hline
    \multirow{10}{*}{\begin{sideways}\small{Chinese}\end{sideways}}
    &MP-LSTM~\cite{venugopalan2014translating} &NAACL2015 &454.0M &0.1819 &0.0906 &0.0511 &0.0315 &0.2211 &0.2246 &0.0630 &- &0.7353 &0.6695 &0.7003 \\
    &TA~\cite{yao2015describing} &ICCV2015  &515.5M               &0.1917 &0.0984 &0.0573 &0.0360 &0.2289 &0.2337 &0.0777 &- &0.7388 &0.6700 &0.7021 \\
    &Transformer~\cite{sharma2018conceptual} &ACL2018  &360.4M    &0.2875 &0.1565 &0.0940 &0.0600 &0.2728 &0.2695 &0.1732 &- &\textbf{0.7529} &0.6998 &0.7248 \\
    &M2NET~\cite{cornia2020meshed} & CVPR2020 &164.8M             &0.3001 &0.1522 &0.0874 &0.0540 &0.2598 &0.2501 &0.1793 &- &0.7283 &0.7047 &0.7158  \\
    &MDVC~\cite{MDVC_Iashin_2020} & CVPRW2020 &325.9M             &0.2918 &0.1466 &0.0823 &0.0503 &0.2566 &0.2520 &0.1537 &- &0.7382 &0.6971 &0.7165  \\
    &XTransformer~\cite{pan2020x} &CVPR2020 &578.2M              &0.2762 &0.1305 &0.0706 &0.0422 &0.2362 &0.2389 &0.1176 &- &0.7215 &0.6890 &0.7044  \\
    &XLAN~\cite{pan2020x} & CVPR2020  &328.2M                     &0.3091 &0.1682 &0.1014 &\textbf{0.0648} &0.2790 &\textbf{0.2766} &0.2077 &- &0.7493 &0.7039 &0.7254  \\
    &BMT~\cite{BMT_Iashin_2020}  & BMVC2020   &340.0M             &0.3216 &0.1621 &0.0912 &0.0559 &0.2614 &0.2563 &0.1784 &- &0.7322 &0.7047 &0.7177  \\
    &MART$^{\ddag}$~\cite{lei2020mart} & ACL2020   &150.6M        &\textbf{0.3601} &\textbf{0.1836} &\textbf{0.1044} &0.0638 &\textbf{0.2815} &0.2725 &\textbf{0.2484} &- &0.7390 &\textbf{0.7148} &\textbf{0.7263}  \\
    &TDPC~\cite{song2021towards} & CVPR2021 &196.6M               &0.3208 &0.1655 &0.0954 &0.0588 &0.2688 &0.2599 &0.1905 &- &0.7357 &0.7112 &0.7227  \\
    \hline
    \hline
    \end{tabular}
    \end{lrbox}
    \scalebox{1.0}{\usebox{\tableboxTabRESEM}}
\end{table*}
\begin{table*}[htbp]
  \caption{Per emotion's statistic performance (METEOR score) of 11 state-of-the-art methods on the proposed EMVPC dataset (English version). The bold font indicates the best performance. MP.=MP-LSTM, Trans.=Transformer, XTrans.=XTransformer. 'all' denotes the METEOR score on the entire dataset.
  }\vspace{-8pt}
  \label{tab:sub-emotion-en}
  \centering
  \renewcommand{\arraystretch}{1.0}
  \renewcommand{\tabcolsep}{5pt}
  \newsavebox{\tableboxTabRESEMEE}
  \begin{lrbox}{\tableboxTabRESEMEE}
  \begin{tabular}{l||cccccccccccc}
  \hline
  \hline
    \multirow{2}{*}{Emotion} &  MP.  & TA & Trans. & UniVL  & M2NET & MDVC & XTrans.  & XLAN  & BMT & MART & TDPC & \multirow{2}{*}{Avg}\\
            &~\cite{venugopalan2014translating} & ~\cite{yao2015describing} & ~\cite{sharma2018conceptual} &  ~\cite{luo2020univl} & ~\cite{cornia2020meshed} &  ~\cite{MDVC_Iashin_2020} &
            ~\cite{pan2020x} &   ~\cite{pan2020x} & ~\cite{BMT_Iashin_2020}  & ~\cite{lei2020mart} &   ~\cite{song2021towards} &  \\
    \hline
    \hline

joy	 &0.3003& 0.3137& 0.3376& 0.3016& 0.3229& 0.3171& 0.3031& \textbf{0.3382}& 0.3196& 0.3244& 0.3206 & 0.3181\\
trust	 &0.2909& 0.3063& \textbf{0.3343}& 0.2859& 0.2929& 0.2780& 0.2819& 0.3172& 0.2886& 0.3007& 0.3100 & 0.2988\\
fear	 &0.2864& 0.3094& \textbf{0.3393}& 0.2922& 0.3049& 0.3027& 0.3019& 0.3302& 0.3131& 0.3134& 0.3230& 0.3106\\
surprise	& 0.2837& 0.3102& 0.3516& 0.3040& 0.3238& 0.3227& 0.3098& \textbf{0.3587}& 0.3149& 0.3376& 0.3327& 0.3227\\
sadness	 &0.2971& 0.3057& 0.3324& 0.3084& 0.3094& 0.3131& 0.3160&\textbf{0.3393}& 0.3117& 0.3288& 0.3243 & 0.3169\\
disgust	 &0.2780& 0.2901& 0.3262& 0.2896& 0.3110& 0.3010& 0.2856&\textbf{0.3390}& 0.3115& 0.3120& 0.3199& 0.3058\\
anger	 &0.2958& 0.3130& 0.3269& 0.3047& 0.3188& 0.3207& 0.3070& 0.3306& 0.3235&\textbf{ 0.3337}& 0.3269& 0.3183\\
anticipation	 &0.2855& 0.3217& 0.3619& 0.3447& \textbf{0.3980}& 0.3278& 0.3019& 0.3925& 0.3285& 0.3766& 0.3326& 0.3429\\
serenity	 &0.3211& 0.3206& 0.3258& 0.2989& 0.3138& 0.3016& 0.3117& \textbf{0.3511}& 0.3059& 0.3058& 0.3028& 0.3145\\
acceptance	& 0.3738& 0.3759& \textbf{0.4189}& 0.3747& 0.3651& 0.3336& 0.3525& 0.4014& 0.3853& 0.3902& 0.3890& 0.3782\\
apprehension	& 0.3221& 0.3303& 0.3412& 0.3039& 0.3305& 0.3352& 0.3191& 0.3457& 0.3111& 0.3253& \textbf{0.3473}& 0.3283\\
distraction	 &0.3156& 0.3194& 0.3185& 0.2960& 0.3119& 0.3181& 0.3002& 0.3300& 0.3210& 0.3000&\textbf{0.3584}& 0.3172\\
pensiveness	& 0.3239& 0.3206& 0.3259& 0.2947& 0.3322& 0.3297& 0.3178& \textbf{0.3437}& 0.3371& 0.3217& 0.3322& 0.3254\\
boredom	 &0.2933& 0.3057& 0.3228& 0.2921& 0.2997& 0.3046& 0.2956& \textbf{0.3259}& 0.3139& 0.3080& 0.3052& 0.3061\\
annoyance	& 0.3238& 0.3097& \textbf{0.3379}& 0.2951& 0.3189& 0.3072& 0.3269& 0.3283& 0.3151& 0.3100& 0.3126 & 0.3169\\
interest	 &0.2559& 0.3128& 0.3235& 0.2821& 0.2458& 0.2884& 0.3089& 0.2753& \textbf{0.3423}& 0.2774& 0.3214& 0.2940\\
ecstasy	 &0.2948& 0.2919& 0.3228& 0.3056& 0.3071& 0.3108& 0.2969& 0.3174& 0.3019& \textbf{0.3248}& 0.3200& 0.3085\\
admiration	& 0.2739& 0.3133& \textbf{0.3609}& 0.3253& 0.3525& 0.3458& 0.3016& 0.3575& 0.3269& 0.3241& 0.3315& 0.3285\\
terror	 &0.2757& 0.3166& 0.2970& 0.2955& 0.2955& 0.2815& 0.2621&\textbf{0.3188}& 0.2789& 0.2871& 0.2825& 0.2901\\
amazement	 &0.3203& 0.3293& \textbf{0.3509}& 0.3254& 0.3147& 0.3320& 0.3135& 0.3429& 0.3345& 0.3240& 0.3309& 0.3289\\
grief	& 0.2940& 0.3035& 0.3183& 0.2878& 0.2987& 0.3133& 0.2976& 0.3183& 0.3001& 0.3007& \textbf{0.3347} & 0.3061\\
loathing	& 0.3113& 0.3330& 0.3423& 0.3153& 0.3071& 0.3310& 0.3347& \textbf{0.3430}& 0.3280& 0.3194& 0.3394& 0.3277\\
rage	& 0.3380& 0.3357&\textbf{0.3562}& 0.3061& 0.3292& 0.3442& 0.3323& 0.3517& 0.3344& 0.3241& 0.3422& 0.3358\\
vigilance	& 0.3195& 0.3333&\textbf{ 0.3345}& 0.2996& 0.3071& 0.3145& 0.3032& 0.3320& 0.3144& 0.3086& 0.3222& 0.3172\\
love	& 0.3201& 0.3259& 0.3445& 0.3144& 0.3190& 0.3225& 0.3228&\textbf{0.3499}& 0.3296& 0.3367& 0.3346& 0.3291\\
submission	& 0.2755& 0.3102& \textbf{0.3560}& 0.3119& 0.2996& 0.3224& 0.3398& 0.3045& 0.2924& 0.3449& 0.3077& 0.3150\\
awe	 &0.3554& 0.3563& \textbf{0.4119}& 0.3861& 0.3696& 0.3754& 0.3787& 0.3975& 0.4009& 0.3645& 0.3461& 0.3766\\
disapproval	& 0.3286& 0.3046& 0.3661& 0.3349& 0.2975& 0.3155& 0.2909& \textbf{0.3686}& 0.3308& 0.3479& 0.3543& 0.3309\\
remorse	 &0.3169& 0.2864& 0.3831& 0.3793& 0.3985& 0.3563& 0.2627& \textbf{0.4059}& 0.3569& 0.3402& 0.3411& 0.3479\\
contempt	 &0.2656& 0.3017& 0.3239& 0.2942& 0.3212& 0.3087& 0.2594& \textbf{0.3407}& 0.3350& 0.3262& 0.3088& 0.3078\\
aggressiveness	& 0.3158& 0.3498& 0.3824& 0.3554& 0.3501& 0.3569& 0.3546& 0.3658& \textbf{0.3868}& 0.3798& 0.3520& 0.3590\\
optimism	& 0.2864& 0.2897&\textbf{0.3297}& 0.2812& 0.3184& 0.2928& 0.2542& 0.3132& 0.3112& 0.3041& 0.3052& 0.2987\\
anxiety	& 0.3064& 0.3097& 0.3469& 0.3070& 0.3374& 0.3258& 0.2840& \textbf{0.3571}& 0.3343& 0.3180& 0.3157& 0.3220\\
pride	& 0.2881& 0.3115& \textbf{0.3495}& 0.2928& 0.2948& 0.2676& 0.3267& 0.3093& 0.3075& 0.2748& 0.3122& 0.3032\\
despair	 &0.3059& 0.3061& \textbf{0.3413}& 0.2980& 0.3083& 0.2860& 0.2948& 0.3402& 0.3315& 0.3091& 0.3040& 0.3114\\
shame	& 0.4108& \textbf{0.5051}& 0.4856& 0.3588& 0.4021& 0.4060& 0.4124& 0.4390& 0.4373& 0.4313& 0.4689& 0.4325\\
guilt	& 0.3140& 0.3234& 0.3405& 0.2953& 0.3209& 0.3405& 0.3232& 0.3412& 0.3246& 0.3217& \textbf{0.3607}& 0.3278\\
curiosity	& 0.2748& 0.2913& 0.3291& 0.2804& 0.2773& 0.2919& 0.2726& \textbf{0.3506}& 0.2994& 0.2903& 0.2723& 0.2936\\
unbelief	& 0.3142& 0.3402& 0.3803& 0.3047& \textbf{0.3931}& 0.3590& 0.3403& 0.3758& 0.3382& 0.3164& 0.3372& 0.3454\\
envy	& 0.2779& 0.3428& 0.3791& 0.2664& 0.3585& 0.3283& 0.3707&\textbf{0.3805}& 0.3100& 0.3016& 0.3501& 0.3333\\
cynicism	& 0.2874& 0.3081&\textbf{0.3581}& 0.3036& 0.2999& 0.2780& 0.3122& 0.3438& 0.2945& 0.3233& 0.3003& 0.3099\\
sentimentality	& 0.2578& 0.2941& 0.2835& 0.2645& 0.2795& 0.2899& 0.2812& 0.3399& \textbf{0.3157}& 0.2807& 0.2974& 0.2895\\
morbidness	& 0.3838& 0.3261& 0.3668& 0.3265& 0.4093& 0.3372& 0.4156& \textbf{0.4296}& 0.2652& 0.3734& 0.3436& 0.3616\\
dominance	& 0.2965& 0.3348& 0.3432& 0.2894& 0.3412& 0.3023& 0.3793& 0.3478& 0.2744& 0.3806& \textbf{0.3890}& 0.3344\\
pain	& 0.3153& 0.3417& 0.3554& 0.3144& 0.3292& 0.3514& 0.3209& \textbf{0.3637}& 0.3115& 0.3304& 0.3329& 0.3333\\
excitement	 &0.3361& 0.3321& \textbf{0.3640}& 0.3263& 0.3045& 0.3494& 0.3431& 0.3506& 0.3461& 0.3390& 0.3354& 0.3388\\
embarrassment	& 0.3124& 0.3285& 0.3508& 0.3151& 0.3176& 0.3423& 0.3088&\textbf{0.3910}& 0.3358& 0.3196& 0.3489& 0.3337\\
nervousness	& 0.3260& 0.3266& 0.3530& 0.3329& 0.3411& \textbf{0.3691}& 0.3179& 0.3599& 0.3248& 0.3479& 0.3378& 0.3397\\
hesitation	& 0.3111& 0.3107& 0.3735& 0.3760& \textbf{0.3921}& 0.3639& 0.3087& 0.3500& 0.3624& 0.3579& 0.3662& 0.3520\\
disappointment	& 0.3378& 0.2854& 0.3610& 0.3401& 0.3455& 0.3290& \textbf{0.4399}& 0.3991& 0.3492& 0.3331& 0.3454& 0.3514\\
confusion	& 0.3495& 0.3431& \textbf{0.3855}& 0.3462& 0.3314& 0.3450& 0.3540& 0.3818& 0.3558& 0.3534& 0.3757& 0.3565\\
sympathy	& 0.2087& 0.1930& 0.2281& 0.2064& 0.2075& 0.1979& 0.2077& \textbf{0.3126}& 0.2025& 0.2239& 0.2303& 0.2199\\
appreciation	& 0.1892& 0.2618& 0.3192& 0.3529& 0.2970& \textbf{0.3609}& 0.2723& 0.2721& 0.3214& 0.2963& 0.3592 & 0.3002\\
\hline
all  & 0.3025 & 0.3161 & 0.3421 & 0.3072 & 0.3202 & 0.3196 & 0.3093 & \textbf{0.3435} & 0.3299 & 0.3273 & 0.3274 &0.3223\\
  \hline
  \hline
  \end{tabular}
  \end{lrbox}
  \scalebox{1.0}{\usebox{\tableboxTabRESEMEE}}
\end{table*}

\begin{table*}[htbp]
  \caption{ Per emotion's statistic performance (METEOR score) of 10 state-of-the-art methods on the proposed EMVPC dataset (Chinese version). The bold font indicates the best performance. MP.=MP-LSTM, Trans.=Transformer, XTrans.=XTransformer. 'all' denotes the METEOR score on the entire dataset.
  }\vspace{-8pt}
  \label{tab:sub-emotion-cn}
  \centering
  \renewcommand{\arraystretch}{1.0}
  \renewcommand{\tabcolsep}{5pt}
  \newsavebox{\tableboxTabRESEMEC}
  \begin{lrbox}{\tableboxTabRESEMEC}
  \begin{tabular}{l||ccccccccccc}
  \hline
  \hline
  \multirow{2}{*}{Emotion}   &  MP.  & TA & Trans. & M2NET & MDVC & XTrans.  & XLAN  & BMT & MART & TDPC &\multirow{2}{*}{Avg}\\
            &~\cite{venugopalan2014translating} & ~\cite{yao2015describing} & ~\cite{sharma2018conceptual}  & ~\cite{cornia2020meshed} &  ~\cite{MDVC_Iashin_2020} &
            ~\cite{pan2020x} &   ~\cite{pan2020x} & ~\cite{BMT_Iashin_2020}  & ~\cite{lei2020mart} &   ~\cite{song2021towards}  & \\
    \hline
    \hline
  joy	 &0.2354& 0.2528& 0.2900& 0.2634& 0.2742& 0.2379& 0.2831& 0.2618& \textbf{0.2923}& 0.2801 & 0.2671\\
trust	& 0.1749& 0.1995& 0.2317& 0.2178& 0.2160& 0.2094& 0.2221& 0.2035& \textbf{0.2326}& 0.2207 & 0.2128\\
fear	 &0.1952& 0.2005& 0.2485& 0.2282& 0.2270& 0.2212& 0.2513& 0.2297& \textbf{0.2519}& 0.2353 & 0.2289\\
surprise& 0.2132& 0.2219& \textbf{0.2810}& 0.2574& 0.2460& 0.2356& 0.2772& 0.2621& 0.2786& 0.2641 & 0.2537\\
sadness	 &0.2022& 0.2026& 0.2549& 0.2444& 0.2497& 0.2160& 0.2546& 0.2580& \textbf{0.2694}& 0.2531 & 0.2405\\
disgust	 &0.1636& 0.1797& 0.2493& 0.2500& 0.2294& 0.2059& \textbf{0.2671}& 0.2592& 0.2655& 0.2645 & 0.2334\\
anger	& 0.2167& 0.2235& 0.2551& 0.2505& 0.2548& 0.2405& 0.2769& 0.2697& \textbf{0.2801}& 0.2614 & 0.2529\\
anticipation& 0.2337& 0.2218& 0.3151& 0.2874& 0.3038& 0.2205& \textbf{0.3386}& 0.2864& 0.3247& 0.2714 & 0.2803\\
serenity	& 0.2208& 0.2203& 0.2753& 0.2717& 0.2551& 0.2335& 0.2668& 0.2504& \textbf{0.2851}& 0.2482 & 0.2527\\
acceptance	& 0.3159& 0.2904& 0.3531& 0.3428& 0.3410& 0.3230& \textbf{0.3651}& 0.2970& 0.3629& 0.3386 & 0.3330\\
apprehension& 0.2228& 0.2390& 0.2600& 0.2528& 0.2552& 0.2556& 0.2878& 0.2646& \textbf{0.2940}& 0.2740 & 0.2606\\
distraction	& 0.2107& 0.1967& 0.2283& 0.2283& 0.2314& 0.2278& 0.2672& 0.2519&\textbf{0.2674}& 0.2423 & 0.2352\\
pensiveness	& 0.2408& 0.2498& 0.2651& 0.2967& 0.2581& 0.2413& 0.2851& 0.2607& \textbf{0.2979}& 0.2772 & 0.2673\\
boredom	    & 0.2010& 0.2044& 0.2384& 0.2335& 0.2203& 0.2169& 0.2360& 0.2315& \textbf{0.2406}& 0.2271 & 0.2250\\
annoyance	& 0.2155& 0.2145& 0.2597& 0.2394& 0.2337& 0.2255& 0.2698& 0.2450& \textbf{0.2818}& 0.2397 & 0.2425\\
interest	& 0.2102& 0.1971&\textbf{ 0.2762}& 0.2489& 0.2095& 0.1976& 0.2738& 0.2451& 0.1979& 0.2341 & 0.2290\\
ecstasy	    & 0.2159& 0.2148& 0.2448& 0.2523& 0.2241& 0.2221& 0.2420& 0.2566& 0.2518& \textbf{0.2636} & 0.2388\\
admiration	& 0.1917& 0.2132& 0.2522& 0.2354& 0.2207& 0.2233& \textbf{0.2938}& 0.2236& 0.2593& 0.2700 & 0.2383\\
terror	    & 0.1613& 0.1922& \textbf{0.2631}& 0.2323& 0.2076& 0.1847& 0.2479& 0.2092& 0.2246& 0.2319 & 0.2155\\
amazement	& 0.2359& 0.2476& 0.2639& 0.2671& 0.2691& 0.2510& 0.2790& 0.2574& \textbf{0.2879}& 0.2752 & 0.2634\\
grief	    &0.1971& 0.1952& 0.2384& 0.2399& 0.2148& 0.1932& \textbf{0.2548}& 0.2139& 0.2384& 0.2504& 0.2236\\
loathing	& 0.2118& 0.2195& 0.2349& \textbf{0.2920}& 0.2283& 0.2319& 0.2780& 0.2914& 0.2649& 0.2636& 0.2516\\
rage	    & 0.2409& 0.2368& \textbf{0.2954}& 0.2712& 0.2659& 0.2631& 0.2855& 0.2852& 0.2879& 0.2901& 0.2722\\
vigilance	& 0.2111& 0.2167& 0.2709& 0.2489& 0.2471& 0.2164& \textbf{0.2937}& 0.2446& 0.2600& 0.2517& 0.2461\\
love	    & 0.2517& 0.2557& 0.2831& 0.2658& 0.2665& 0.2593& 0.2857& 0.2583&\textbf{0.2940}& 0.2801& 0.2700\\
submission	& 0.2003& 0.2292& 0.2404& 0.2569& 0.2431& 0.2439& 0.2492& 0.2366&\textbf{ 0.2634}& 0.2501& 0.2413\\
awe	        & 0.4200& \textbf{0.4233}& 0.3630& 0.3201& 0.3928& 0.3198& 0.3997& 0.3927& 0.4018& 0.4210& 0.3854\\
disapproval	& 0.2439& 0.2474& 0.2892& 0.2933& 0.2574& 0.2599& \textbf{0.3099}& 0.2804& 0.3089& 0.3037& 0.2794\\
remorse	    & 0.1833& 0.2652& 0.2828& 0.2806& 0.2173& 0.1773& \textbf{0.3325}& 0.2597& 0.2755& 0.2443& 0.2519\\
contempt	& 0.2057& 0.2072& 0.2910& 0.2559& 0.2513& 0.2059& \textbf{0.3151}& 0.2918& 0.2869& 0.2781& 0.2589\\
aggressiveness	& 0.2722& 0.2867& 0.3350& 0.3305& 0.3448& 0.2535& 0.3577& 0.3661& \textbf{0.3704}& 0.3365& 0.3253\\
optimism	    & 0.1854& 0.2037& \textbf{0.2664}& 0.2440& 0.2241& 0.2041& 0.2347& 0.2369& 0.2467& 0.2553& 0.2301\\
anxiety	        & 0.1991& 0.2528& 0.3113& 0.3083& 0.2770& 0.2595& \textbf{0.3160}& 0.2828& 0.3127& 0.3027& 0.2822\\
pride	        & 0.2423& 0.2219& 0.2284& \textbf{0.2673}& 0.2318& 0.2322& 0.2414& 0.2399& 0.2613& 0.2513& 0.2418\\
despair	        & 0.1840& 0.1965& 0.2359& 0.2254& 0.2261& 0.2057& 0.2383& 0.2106& \textbf{0.2675}& 0.2410& 0.2231\\
shame	        & 0.4075& 0.4306& \textbf{0.5226}& 0.4009& 0.4812& 0.4720& 0.4501& 0.4858& 0.4808& 0.4853& 0.4617\\
guilt	        & 0.2445& 0.2245& 0.2785& \textbf{0.2808}& 0.2062& 0.2726& 0.2592& 0.2619& 0.2685& 0.2794& 0.2576\\
curiosity	& 0.1843& 0.1969& 0.2495& 0.2396& 0.2334& 0.2047& 0.2559& 0.1986& \textbf{0.2561}& 0.2315& 0.2251\\
unbelief	& 0.1970& 0.2106& 0.2889& 0.2821& 0.2705& 0.2322& 0.2825& 0.2752& \textbf{0.3106}& 0.2976& 0.2647\\
envy	    & 0.1875& 0.1788& 0.2322& 0.2365& \textbf{0.3374}& 0.1763& 0.3071& 0.2567& 0.2659& 0.2470& 0.2425\\
cynicism	& 0.2264& 0.2097& 0.2577& 0.2424& 0.2372& 0.2630& \textbf{0.2772}& 0.2267& 0.2471& 0.2372& 0.2425\\
sentimentality	& 0.1557& 0.1481& 0.1763& 0.1836& 0.1641& 0.1690& 0.1951& 0.1529& 0.1453& \textbf{0.2047}& 0.1695\\
morbidness	& 0.1962& 0.2191& \textbf{0.3806}& 0.2478& 0.3213& 0.3049& 0.2974& 0.3066& 0.3672& 0.2753& 0.2916\\
dominance	& 0.2118& 0.1957& 0.2478& 0.2525& 0.2235& 0.1824& \textbf{0.3128}& 0.2035& 0.2398& 0.2736& 0.2343\\
pain	    & 0.2467& 0.2377& 0.2904& 0.2704& 0.2752& 0.2515& \textbf{0.2985}& 0.2664& 0.2699& 0.2805& 0.2687\\
excitement	& 0.2641& 0.2627& 0.2913& 0.2655& 0.2747& 0.2528& \textbf{0.3011}& 0.2660& 0.2882& 0.2879& 0.2754\\
embarrassment	& 0.2662& 0.2670& 0.2799& 0.2757& 0.2910& 0.2848& \textbf{0.3002}& 0.2578& 0.2832& 0.2801& 0.2786\\
nervousness	& 0.2615& 0.2446& 0.2725& 0.2497& 0.2722& 0.2446& 0.2759& 0.2598& \textbf{0.2893}& 0.2783& 0.2648\\
hesitation	& 0.2111& 0.1913& 0.3385& 0.3343& 0.3014& 0.2289& 0.2911& 0.2856& \textbf{0.3546}& 0.3021& 0.2839\\
disappointment	& 0.1711& 0.1663& 0.2192& 0.2178& \textbf{0.2974}& 0.2347& 0.2837& 0.2542& 0.2607& 0.2236& 0.2329\\
confusion	& 0.2918& 0.2647& \textbf{0.3004}& 0.2458& 0.2660& 0.2882& 0.2572& 0.2724& 0.2981& 0.2916& 0.2776\\
sympathy	& 0.1817& 0.1924& 0.1615& 0.2074& 0.2634& 0.1963& \textbf{0.3479}& 0.2232& 0.2254& 0.1887& 0.2188\\
appreciation	& 0.1653& 0.1367& 0.1674& 0.1912& 0.1800& 0.1495& 0.2452& 0.2239& \textbf{0.2497}& 0.1902& 0.1899\\
\hline
all & 0.2211 & 0.2289 & 0.2728 & 0.2598 & 0.2566 & 0.2362 & 0.2790 & 0.2614 & \textbf{0.2815} & 0.2688 & 0.2566\\
    \hline
    \hline
    \end{tabular}
  \end{lrbox}
  \scalebox{1.0}{\usebox{\tableboxTabRESEMEC}}
\end{table*}
\subsection{Quantitative Comparison}

\subsubsection{Performance on Charades Captions}

The Charades Captions dataset~\cite{wang2018video} usually has 1-2 descriptive sentences for each video. The benchmark results of existing state-of-the-art approaches on Charades Captions are represented in Table~\ref{tab:comparision-anet-charades}, where it is observed that the Transformer~\cite{sharma2018conceptual} and UniVL~\cite{luo2020univl} models obtain the top-2 performance. Interestingly, the image captioning method Transformer achieves the best BLEU-1, BLEU-2, CIDEr, SPICE and $F_{bert}$ scores of 0.6522, 0.4513, 0.2942, 0.1737 and 0.8990, which even outperforms the latest video paragraph captioning method TDPC~\cite{song2021towards}. Moreover, the methods MP-LSTM~\cite{venugopalan2014translating} and TA~\cite{yao2015describing} developed in early years get promising statistic results compared with their performances on ActivityNet Captions and EMVPC. The experiment implies that the image captioning and video captioning methods are more suitable to generate satisfying captions on short videos.

\subsubsection{Performance on ActivityNet Captions}

The ActivityNet Captions dataset~\cite{krishna2017dense} is the most popular and largest dataset for video paragraph captioning. The comparison in Table~\ref{tab:comparision-anet-charades} shows that many methods obtain competitive performance on this dataset, \emph{e.g.}, XLAN~\cite{pan2020x} achieves the best BLEU-2, BLEU-3 and BLEU-4 scores of 0.3157, 0.2034 and 0.1338, MART~\cite{lei2020mart} achieves the best BLEU-1, METEOR and SPICE scores of 0.5198, 0.3544 and 0.1710, TDPC~\cite{song2021towards} achieves the best CIDEr, $P_{bert}$ and $F_{bert}$ scores of 0.2737, 0.8890 and 0.8833, Transformer~\cite{sharma2018conceptual} achieves the best ROUGE\_L score of 0.3459, and AdvInf$^{\ddag}$~\cite{park2019adversarial} achieves the best $R_{bert}$ score of 0.8784. Interestingly, it can be observed that no method obtains the best performance in most metrics and the methods with annotations (\textit{e.g.}, MART) do not achieve absolute advantages than those without annotations.

\subsubsection{Performance on EMVPC}

The overall statistic comparisons on the proposed EMVPC are reported in Table~\ref{tab:comparision}. The top-3 performance methods on English version are XLAN~\cite{pan2020x}, Transformer~\cite{sharma2018conceptual} and BMT~\cite{BMT_Iashin_2020}, their BLEU-4 scores are 0.0899, 0.0873, 0.0803 and METEOR scores are 0.3405, 0.3421, 0.3299. For the Chinese version of EMVPC, MART~\cite{lei2020mart} obtains the best statistic performance in most metrics, where the BLEU-1, BLEU-2, BLEU-3, METEOR, CIDEr, $R_{bert}$ and $F_{bert}$ scores are 0.3601, 0.1836, 0.1044, 0.2815, 0.2484, 0.7148 and 0.7263, respectively. Nonetheless, it can be observed that the models show worse performance compared with the statistic results on the other two datasets. The unsatisfied results demonstrate the challenges of EMVPC and more effective models need to be designed for performance improvement. Note that some image captioning methods (\textit{e.g.}, XLAN~\cite{pan2020x}) can even achieve better performance than the video paragraph captioning models (\textit{e.g.}, TDPC~\cite{song2021towards}), demonstrating the potential capacity to leverage image captioning based technology to address the video paragraph captioning task. Another interesting observation is that the evaluation results on Chinese version perform worse than that on English version, for example, the BLEU-4 and METEOR scores of BMT~\cite{BMT_Iashin_2020} are 0.0803 and 0.3299 on English version, while the scores clearly drop to 0.0559 and 0.2614 on Chinese version. The reasons may lie in the following two aspects: (1) inaccurate predictions caused by the long-tail distribution of larger Chinese vocabulary, and (2) more complex Chinese syntax and grammatical structures, making it difficult to learn coherent, accurate and expressive sentences.

In order to provide more detailed analysis about each emotion, the evaluation results about 53 emotions are further reported in Table~\ref{tab:sub-emotion-en} and Table~\ref{tab:sub-emotion-cn}. It can be observed that lower METEOR scores are obtained in the emotions such as terror, curiosity, sentimentality and sympathy, which contain abstract or implicit emotions in untrimmed videos. Moreover, XLAN and Transformer are the top-2 performance models on English version, which obtain the best METEOR scores in 20 and 16 emotions, respectively. MART achieves the best METEOR scores in most emotion categories on Chinese version.

Moreover, the accuracy of emotion words and emotion sentences of the generated paragraphs on EMVPC is further evaluated by employing the metrics of {ACC$_{sw}$}, {ACC$_{c}$}, BFS and CFS~\cite{wang2021emotion}, with the comparison results shown in Table~\ref{tab:emotion score}. From the results, it can be observed that XLAN and Transformer achieve the top-2 statistic results, which outperform other methods by a large margin. Nonetheless, the emotion scores of the competing methods are unsatisfying and need to be improved in future works.

\begin{table}[htbp]
  \caption{The emotion accuracy of existing state-of-the-art methods on EMVPC. $\ddag$ denotes the model using segment annotations in both training and testing procedures, - means the statistic results are not available.}\vspace{-8pt}
  \label{tab:emotion score}
  \centering
  \renewcommand{\arraystretch}{1.1}
  \renewcommand{\tabcolsep}{1pt}
  \newsavebox{\tableboxTabEMNUM}
  \begin{lrbox}{\tableboxTabEMNUM}
  \begin{tabular}{l||cccc||cccc}
  \hline
  \hline
  Datasets & \multicolumn{4}{c||}{EMVPC (English)} & \multicolumn{4}{c}{EMVPC (Chinese)} \\
  \hline
  \hline
    Methods & ACC$_{sw}$ & ACC$_{c}$ & BFS & CFS & ACC$_{sw}$ & ACC$_{c}$ & BFS & CFS \\
    \hline
    \hline
    MP-LSTM~\cite{venugopalan2014translating} & 22.64& 17.68& 11.61& 9.14     & 15.77& 7.92& 7.51& 7.41  \\
    TA~\cite{yao2015describing} & 25.31& 20.69& 13.15& 10.74      & 16.69& 10.27& 8.33& 8.91  \\
    Transformer~\cite{sharma2018conceptual} & 37.43& 29.53& 19.13& 19.10   & 35.84& \textbf{21.31}& 14.70& 19.57  \\
    UniVL~\cite{luo2020univl} &  33.80& 28.46& 16.79& 17.95    &    -& -& -& -  \\
    M2NET~\cite{cornia2020meshed} & 32.61& 10.70& 14.99& 16.48      & 28.75& 5.25& 12.06& 17.74    \\
    MDVC~\cite{MDVC_Iashin_2020} & 30.75& 26.03& 17.77& 16.85    & 27.75& 10.02& 12.04& 16.07    \\
    XTransformer~\cite{pan2020x} & 26.81& 26.52& 16.08& 14.39      & 19.24& 6.38& 9.90& 11.97    \\
    XLAN~\cite{pan2020x} & \textbf{37.91}& \textbf{32.49}& \textbf{19.54}& 21.02       & \textbf{37.65}& 20.88& \textbf{15.52}& 22.47     \\
    BMT~\cite{BMT_Iashin_2020} & 29.85& 9.24& 16.21& 16.01    & 26.98& 5.92& 12.43& 17.56  \\
    MART$^{\ddag}$~\cite{lei2020mart}& 33.98& 28.75& 17.68& \textbf{25.05}     & 30.70& 8.07& 14.24& \textbf{23.75}   \\
    TDPC~\cite{song2021towards} & 29.19& 26.86& 17.61& 16.83      & 26.03& 4.64& 12.45& 18.31  \\
    \hline
    \hline
    \end{tabular}
    \end{lrbox}
    \scalebox{0.95}{\usebox{\tableboxTabEMNUM}}
\end{table}

\section{Discussion and Future Direction}
\label{sect:discs}

From the quantitative and qualitative evaluations in Section~\ref{sect:expmt}, it is obvious that the image captioning methods (\textit{e.g.}, M2NET~\cite{cornia2020meshed}, XLAN~\cite{pan2020x}) or dense video captioning methods (\textit{e.g.}, MDVC~\cite{MDVC_Iashin_2020}, BMT~\cite{BMT_Iashin_2020}) can be easily transformed to address the video paragraph captioning task and have achieved promising performances. Nonetheless, the statistic results on video paragraph captioning datasets are worse than those on other visual captioning datasets. For example, the ROUGE\_L scores of XLAN on the MS COCO~\cite{lin2014microsoft} dataset and the proposed EMVPC (English version) dataset are 0.5800 and 0.3084, respectively. A number of challenging problems in this area lead to the ineffectiveness of models and still need to be solved. In order to find out the weaknesses of existing methods and solve these problems in future researches, four important issues are discussed in the following to guide the direction of improving video paragraph captioning.

\subsection{Efficiency}

Efficiency is regarded as one of the most important standards when constructing the framework of captioning model, which implies a model's real-time performance to handle dense and complex videos. It is acknowledged that the duration of the input video is long and the generated paragraph usually contains multiple events, so the methods which do not take the efficiency into consideration will increase the computation cost. Therefore, solving the efficiency problem will boost the deployment of video paragraph captioning models to practical applications.

\subsection{Topicality}

Topicality reflects the capability to capture multiple happened events and identify the accurate topic from video contents, facilitating to the generation of final descriptions. However, many existing models cannot summarize multiple events well and incorporate the most important emotion into the generated paragraph. This influences the generalization of topicality and will also damage the performance of video paragraph captioning approaches.

\subsection{Logicality}

It is an important yet difficult task to ensure reasonable associations between generated sentences. Most advanced captioning methods only focus on discovering local connections instead of considering non-local contextual information of extracted multi-modal features, leading to produce incoherent and unreasonable sentences. To solve such a challenge, designing a framework with logicality should be considered, for generating fluent and logic expressions easily.

\subsection{Metric}

The evaluation metrics in image captioning and text generation areas can also be applied to video paragraph captioning, \textit{i.e.}, measuring the n-gram overlapping or calculating contextual embeddings' similarity between candidate and reference. However, the generated paragraph contains logical relationships among sentences and topic-aware emotions, which is more comprehensive and increases the evaluation difficulty. Moreover, traditional metrics are suitable to evaluate single sentence, rather than judging the potential relationships in multiple sentences. Therefore, more suitable metrics need to be designed in future researches.

\section{Conclusion}
\label{sect:conln}

In this work, a comprehensive investigation is presented for video paragraph captioning. Upon analyzing the problem of insufficient learning data for video paragraph captioning, \textit{e.g.}, lacking of logical association and emotional indications, a large-scale emotion and logic driven multilingual dataset named EMVPC is developed in order to provide a deeper description of video contents. In particular, this EMVPC dataset consists of 10,291 high-quality videos and 20,582 elaborate paragraphs with English and Chinese captions, with large number of human emotion indications, logic of annotated paragraphs and narrations of complex plots. We believe this dataset is a paradigm shifting milestone in terms of rich emotions, context logic and detailed expressions that promote the full-fledge development of new research related to vision-language tasks. Besides, this paper also reports a comprehensive and fundamental study based on experiments conducted on three benchmark datasets (Charades Captions, ActivityNet Captions and the proposed EMVPC). The results based on existing state-of-the-art approaches from different visual captioning tasks are evaluated extensively in terms of 15 popular metrics. Finally, based on the experimental results, the existing challenges are analyzed and insightful suggestions for future directions are also povided. It is expected that this work shall facilitate further advances in video paragraph captioning research.

{\small
\bibliographystyle{IEEEtran}
\bibliography{egbib}
}




\vfill


\end{document}